\documentclass{article}
\usepackage{graphicx}
\usepackage[main, final]{neurips_2026}

\usepackage[utf8]{inputenc} % allow utf-8 input
\usepackage[T1]{fontenc}    % use 8-bit T1 fonts
\usepackage{hyperref}       % hyperlinks
\usepackage{url}            % simple URL typesetting
\usepackage{booktabs}       % professional-quality tables
\usepackage{amsfonts}  
\usepackage{amsmath}% blackboard math symbols

\usepackage{nicefrac}       % compact symbols for 1/2, etc.
\usepackage{microtype}      % microtypography
\usepackage{xcolor}         % colors
\usepackage{kotex}
\usepackage{booktabs}
\usepackage{makecell}
\usepackage{graphicx}
\usepackage{subcaption}
\usepackage{multirow}
\usepackage{array}
\usepackage[normalem]{ulem}
\usepackage{wrapfig}
\usepackage{svg}
\usepackage{placeins}
\usepackage[table]{xcolor}
\usepackage{ulem}

\title{Rethinking LLM Fine-Tuning via Weight Space Reparameterization: Preserving Safety during Downstream Adaptation}

\author{%
  Min-Seong Kim\textsuperscript{*} \quad
  Jongbok Won\textsuperscript{*} \quad
  Jeesup Park \quad
  Yonghee Choi \quad
  Dong-Jun Han\\[4pt]%\textsuperscript{\dag} \\[4pt]
  Yonsei University \\[4pt]
  \texttt{\{minseongkim, jongbok.won, jeesup, 2025321337, djh\}@yonsei.ac.kr}
}

\begin{document}

\maketitle
\begingroup
\renewcommand{\thefootnote}{}
\footnotetext{\textsuperscript{*}Equal contribution. %\textsuperscript{\dag}Corresponding author.
}
\footnotetext{Code available at \url{https://github.com/min-seong-kim/WSR-Tune}}
\endgroup
\begin{abstract}
Fine-tuning large language models (LLMs) on downstream tasks often weakens their previously aligned safety behavior, creating an inherent trade-off  between task adaptation and safety preservation. Existing approaches attempt to mitigate this issue by identifying safety-relevant neurons, layers, or update directions in the original parameter space. However, safety-related information is often distributed across many directions in this space, causing downstream updates to overwrite parameters that encode safety behavior. 
We propose \textbf{WSR-Tune}, a \uline{W}eight \uline{S}pace \uline{R}eparameterization-based fine-tuning framework that preserves safety during downstream adaptation. Our approach constructs a safety-conditioned basis and reparameterizes the weight matrices such that safety-relevant information becomes concentrated in a small subset of directions. We then identify safety-critical directions and freeze them in the reparameterized space, while updating the remaining complementary directions for downstream adaptation. By explicitly structuring the parameter space to disentangle safety and task information, our method reduces interference between objectives and enables simultaneous preservation of safety and improvement in downstream performance. %We further extend this reparameterization principle to parameter-efficient fine-tuning through \textbf{WSR-LoRA}. 
 Experimental results show that our method achieves a more favorable trade-off between downstream performance and safety retention, demonstrating its effectiveness for reliable LLM fine-tuning. 
\end{abstract}

\section{Introduction}
\label{sec:introduction}

% \wjb{ref.bib 미수정} \wjb{ActSVD 관련한 서술 고민: 포함할 것인가? Appendix 부분에서도 굳이 보여줄 필요가 있나 싶은 생각이라, discussion section은 남겨두되, ActSVD의 표현 자체를 사용하지 않는게 어떨까 싶어서요.} \kms{약점 방어용으로 적은 느낌이 어느정도 있어서 제거하는 편이 나을 것 같습니다. 굳이 related work에서 이후 appendix의 detailed dicsuccion을 보여준다고도 하지말고 appendix도 가리는 편이 좋아보입니다.} \wjb{저는 현재 내용들은 다 넣되, ActSVD와 같은 특정 method에 대한 방어적 표현만 가리는 편이 좋지 않나 생각이 드는 것 같습니다.. 가령 Appendix I의 경우에는, paragraph 명을 ActSVD로 사용하지 않고, Original Space에서의 null space projection 기법과 비교하는 느낌으로만 서술해도 되지 않나 싶은 생각입니다. (citation 을 actsvd로 달면 어떨까 싶습니다.)}
% \wjb{Appendix 에서의 paragraph를 textbf로 바꾸는게 낫지 않을까 싶습니다.} \wjb{Appendix Table Design 관련 comment 확인 부탁드립니다.}\wjb{ICLR 버전에서 내렸던 table을 하나 정도 올리는 편이 좋지 않나 싶습니다.}

Large Language Models (LLMs) are increasingly being deployed in real-world settings where safety is critical~\citep{trustllm}. As their capabilities continue to advance, LLMs are expected to play important roles across diverse domains such as healthcare, education, and software engineering~\citep{bloomberggptl, chatgpt4good, gptcode, swebench}. Without adequate safety tuning, LLMs can provide harmful assistance, violate policies, and exhibit unsafe behavior at scale~\citep{chan0park}. Ensuring that these models remain helpful while consistently refusing harmful or disallowed requests has therefore become a central challenge in modern LLM development.

In practice, LLM deployment typically involves multiple post-training stages: a base model is first improved through instruction tuning and/or safety tuning, and is then further adapted to user-specific downstream tasks~\citep{ouyang2022training}. However, such downstream adaptation can degrade previously learned safety behaviors~\citep{lisa}.
This issue becomes more critical as attempts to misuse LLMs have become increasingly sophisticated~\citep{yu2025infecting, li2025commercialllmagentsvulnerable, yang2024watch}. Recent work has introduced a wide range of jailbreak strategies, including manually crafted prompts~\citep{manual_prompt}, automated prompt optimization~\citep{advbench, autodan}, adversarial fine-tuning~\citep{adv_finetuning, shadow_finetuning}, and other adaptive attacks~\citep{pair, pap, liar} designed to elicit harmful or policy-violating outputs even from safety-tuned models~\citep{jailbroken}. These threats suggest that safety must be preserved throughout model adaptation, deployment, and continual improvement.

% ORIGINAL
% A central challenge in LLM fine-tuning is thus preserving safety during downstream adaptation. To mitigate this issue, a growing body of work has attempted to explicitly preserve safety throughout downstream fine-tuning \citep{safeInstr,sn-tune,safe-lora,safe-delta,Resta}. The authors of \citep{safeInstr} show that adding a small amount of safety data during fine-tuning can provide  safety gains, although excessive safety tuning may induce over-refusal. In \citep{sn-tune}, the authors propose SN-Tune and RSN-Tune, which identify safety-specific neurons and selectively update them to improve safety during downstream adaptation. Other approaches instead modify post-fine-tuning updates: Safe Delta \citep{safe-delta} adjusts delta parameters based on their estimated safety degradation and utility contribution, and applies a safety compensation vector.  Resta \citep{Resta} restores safety by adding a safety vector to the compromised model weights. These methods show that safety can be partially preserved or recovered during downstream adaptation.

%\kms{기존 논문에서는 baseline 기법들이 많이 없어서 나열하면서 설명했지만 추가된 baseline까지 모두 추가하게 되면 너무 분량이 많아지고 이미 같은 내용을 related work에서 설명하여 safety method의 종류가 무엇이 있는지 나열 후 이후 related workd에서 disccusion하겠다고 언급했습니다.}

A central challenge in LLM fine-tuning is thus preserving safety during downstream adaptation, motivating a growing body of safety-preserving methods. \emph{Data- and optimization-based methods} incorporate safety supervision or select safety-compatible training data to guide downstream optimization~\citep{safeInstr,lisa,seal}. \emph{Parameter- and update-constrained methods} protect safety-relevant components or restrict update directions to limit safety degradation~\citep{sn-tune,asft,yi2026safegrad}. \emph{Low-rank and post-hoc model-space methods} constrain low-rank updates or modify fine-tuned weights to preserve or restore safety~\citep{safe-lora,salora,safe-delta,Resta}. These methods show that safety can be partially preserved or recovered during downstream adaptation.%These approaches can partially preserve or recover safety, as discussed in Section~\ref{sec:related}.}

\textbf{Challenges.} However, despite prior efforts, existing methods share a key limitation: they operate entirely in the \textit{original parameterization} of the model. For example, consider a safety-tuned $2 \times 2$ weight matrix $W$ from an arbitrary layer, which can be expressed as
\begin{equation}\label{eq:11}
W = \begin{bmatrix} w_{11} & w_{12} \\ w_{21} & w_{22} \end{bmatrix}= w_{11} b_{11}  + w_{12} b_{12} +  w_{21} b_{21} + w_{22} b_{22},
\end{equation}
where $\{w_{ij}\}$ are the model parameters, $\{b_{ij}\}$ are the canonical basis matrices, and each $b_{ij}$ has a single nonzero entry of 1 at position $(i,j)$ and zeros elsewhere (e.g., {\scriptsize $b_{11} = \begin{bmatrix} 1 & 0 \\ 0 & 0 \end{bmatrix}$}, {\scriptsize $b_{22} = \begin{bmatrix} 0 & 0 \\ 0 & 1 \end{bmatrix}$}).

In this representation, safety-related information is not localized to a specific coordinate $w_{ij}$, but is instead distributed across multiple directions in the parameter space (i.e., across  $w_{11}$, $w_{12}$, $w_{21}$, $w_{22}$). Consequently, even if prior methods identify and preserve safety-relevant parameters, downstream adaptation can still overwrite other parameters that encode safety behavior.  In other words, preserving a subset of parameters in the original space is often insufficient, as the parameterization does not provide a clear separation between safety-preservation and task adaptation.

% Llama-2-7b-base model
\begin{wraptable}{r}{0.42\textwidth}
\vspace{-1.3\baselineskip}
\centering
\small
\caption{Impact of freezing safety-relevant parameters on safety preservation after downstream fine-tuning. 
The reported values are averaged over Direct, AutoDAN, PAIR, and PAP attacks using safety-tuned Llama-2-7B.}
\vspace{-0.5em}
\label{tab:basis_rotation_motivation}
\begin{tabular}{lc}
\toprule
\textbf{Fine-tuning setting} & \textbf{Safety $\downarrow$ } \\
\midrule
%Before downstream FT      & 0.1022 \\
% Before FT     & 52.50 \\
FT (0\% frozen)     & 53.84 \\
FT (10\% frozen)         & 33.95 \\
% FT (20\% frozen)         & 14.97\\
FT (30\% frozen)         & 21.44\\
% FT (40\% frozen)         & 13.51\\
FT (50\% frozen)         & 18.43 \\
%FT (90\% frozen)         & 0.0991 \\
WSR-Tune (10\% frozen)   & \textbf{14.67}\\
\bottomrule
\end{tabular}
\vspace{-1.4\baselineskip}
\end{wraptable}

% % Llama-2-7b-chat model
% \begin{wraptable}{r}{0.42\textwidth}
% \vspace{-1.3\baselineskip}
% \centering
% \small
% \caption{Impact of freezing safety-relevant parameters on safety preservation after downstream fine-tuning. 
% The values are averaged over Direct, AutoDAN, PAIR, and PAP attacks using Llama-2-7B-Chat.}
% \vspace{-0.5em}
% \label{tab:basis_rotation_motivation}
% \begin{tabular}{lc}
% \toprule
% \textbf{Fine-tuning setting} & \textbf{Safety $\downarrow$ } \\
% \midrule
% Before FT     & 5.68 \\
% % After FT (No frozen)     & 20.78 \\
% FT (10\% frozen)         & 9.17 \\
% FT (20\% frozen)         & 7.70\\
% % FT (30\% frozen)         & 6.94\\
% FT (50\% frozen)         & 6.92 \\
% WSR-Tune (10\% frozen)   & \textbf{6.90}\\
% \bottomrule
% \end{tabular}
% \vspace{-1.4\baselineskip}
% \end{wraptable}

Table~\ref{tab:basis_rotation_motivation} empirically demonstrates this limitation. Full-parameter adaptation on a downstream task severely degrades safety, while freezing 10\%, 30\%, and 50\% of the original-space parameters only partially alleviates this degradation.
In contrast, applying the same freeze-and-adapt procedure after reparameterization preserves safety much more effectively while maintaining sufficient capacity for downstream adaptation. These results suggest that robust safety preservation cannot be achieved solely by selecting parameters to freeze in the original space; rather, it requires reparameterizing the model so that safety-preserving and task-adaptive directions are more easily disentangled.
We thus pose the following key question: 

\textit{Can we construct a weight-space representation that concentrates safety-relevant information into a compact subspace, while leaving the remaining directions available for downstream adaptation?}

\textbf{Contributions.} Motivated by this question, we propose \textbf{WSR-Tune}, a safety-aware LLM fine-tuning framework based on \textit{weight-space reparameterization}.  Specifically, we replace the above canonical parameterization in Equation (\ref{eq:11}) with a reparameterized form as
\begin{equation}
W = \begin{bmatrix} w_{11} & w_{12} \\ w_{21} & w_{22} \end{bmatrix} = \tilde{w}_{11} \phi_{11}  + \tilde{w}_{12} \phi_{12} +  \tilde{w}_{21} \phi_{21} + \tilde{w}_{22} \phi_{22},
\end{equation}
where ${\phi_{ij}}$ indicates a new basis and ${\tilde{w}_{ij}}$ are the corresponding weights in the reparameterized space. Unlike the canonical basis where each $b_{ij}$ contains a single nonzero entry, the new basis $\{\phi_{ij}\}$ can consist of directions with multiple nonzero entries.

% \wjb{distributed되어 있던 safety relevant attribute을 reparameterize를 통해 localize한 뒤, localize된 parameter에 대한 functioning(?)을 통해 안전성을 보존하고, 이를 original space로 invert함으로써, 안전 정렬 효과를 (1) localized attribute에 대해서만 적용하되, 그 효과가 (2) original space의 전역에 걸쳐 작용하도록 한 점이라는 점을 novelty로 appeal하면 어떨까 싶음.}

\begin{figure}[t]
   % \vspace{-6mm}
    \centering
    \includegraphics[width=\textwidth]{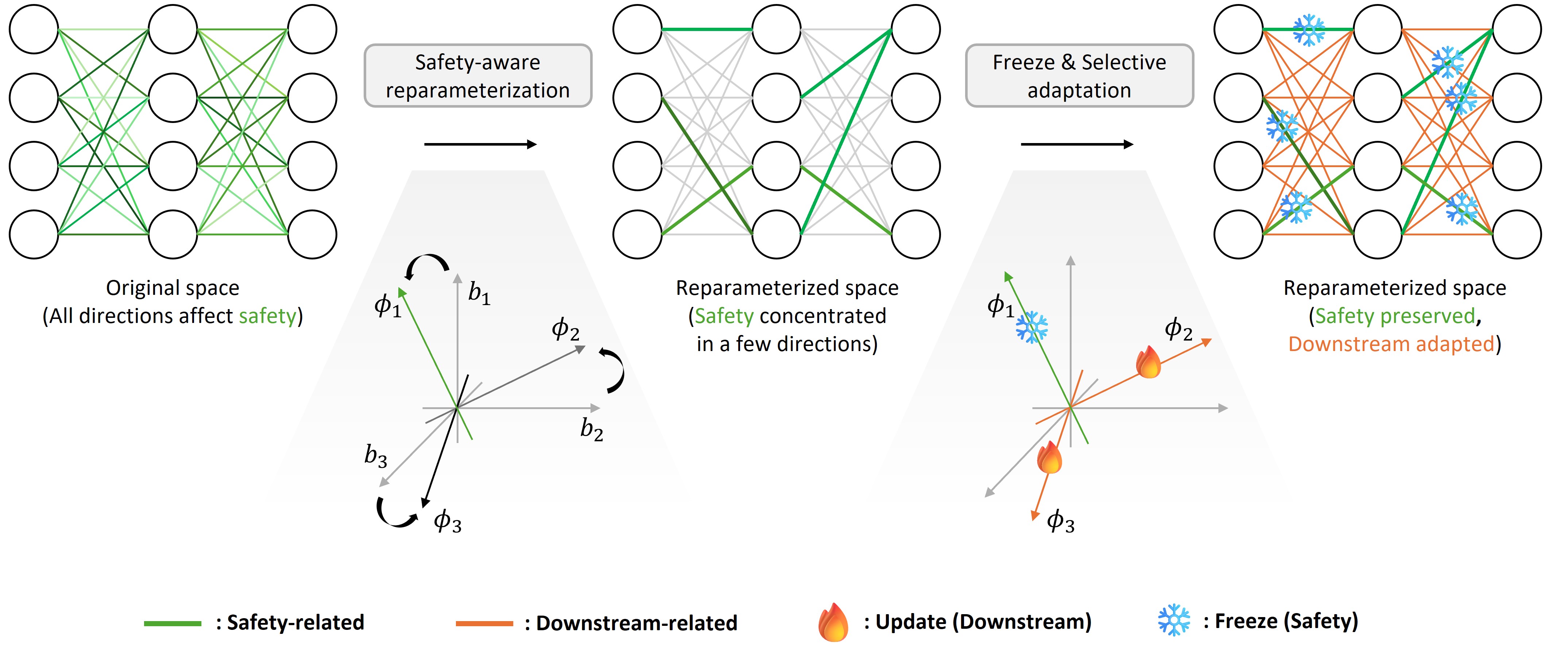}
        \vspace{-4mm}
    \caption{Overview of \textbf{WSR-Tune}. The weight space is reparameterized into a safety-conditioned basis, in which safety-relevant information is concentrated in a subset of coefficients. Safety-critical coefficients are then identified and frozen, while the remaining coefficients are updated for downstream adaptation. This reduces interference between safety and downstream task information during fine-tuning, improving task performance with minimal safety degradation.
    } 
    \label{fig:WSR-Tune_figure}
    \vspace{-2mm}
\end{figure}

Building on this framework, we construct a safety-conditioned basis $\{\phi_{ij}\}$ and reparameterize the weight space from $\{{w}_{ij}\}$ to $\{{\tilde{w}_{ij}}\}$  such that safety-relevant behavior is concentrated in a small subset of directions (i.e., only a few basis directions in $\{\phi_{ij}\}$ capture safety-relevant information), rather than being diffusely distributed.  This safety-conditioned reparameterization is our key contribution: it restructures the weight space to better separate safety-preserving and task-adaptive directions. Under this reparameterized space, we selectively freeze the coefficients associated with safety-critical directions while updating the complementary directions for downstream adaptation, as illustrated in Figure~\ref{fig:WSR-Tune_figure}. As a result, our approach reduces interference between previously learned safety behavior and downstream adaptation, enabling more effective safety-preserving fine-tuning.
% Building on this framework, we construct a safety-conditioned basis $\{\phi_{ij}\}$ and reparameterize the weight space from $\{{w}_{ij}\}$ to $\{{\tilde{w}_{ij}}\}$  such that safety-relevant behavior is concentrated in a small subset of directions (i.e., only a few basis directions in $\{\phi_{ij}\}$ capture safety-relevant information), rather than being diffusely distributed. 
% Our key contribution is a safety-conditioned reparameterization that restructures the weight space to better separate safety-preserving and task-adaptive directions by concentrating safety-critical information into a small subset of coefficients. Under this reparameterized space, we selectively freeze these safety-critical coefficients while updating the complementary directions for downstream adaptation, as illustrated in Figure~\ref{fig:WSR-Tune_figure}. As a result, our approach reduces interference between previously learned safety behavior and downstream adaptation, enabling more effective safety-preserving fine-tuning.

Our WSR-Tune can be seamlessly integrated with existing downstream adaptation methods originally designed for the original parameter space, by applying them in the reparameterized space without modifying their update rules. Experimental results demonstrate that our method better preserves safety during downstream fine-tuning and achieves a more favorable trade-off between downstream performance and safety retention, compared to methods operating in the original parameter space. 

\vspace{-1.5mm}

\section{Related Works}
\label{sec:related}
\vspace{-1.5mm}

% \kms{Reviewer 2가 언급한 reference 논문들 중에서 따로 paragraph을 만들어서 넣을 내용들이 애매합니다. (대부분 safety 성능 향상을 위해 freezing, constraints, safety data selection, 혹은 harmful data가 포함된 malicious train 환경에서 safety robustness) 그래서 기존 paragraph에 대해 몇몇 논문들을 더 추가해주는 방향으로 업데이트했습니다.}

\textbf{Preserving safety during downstream adaptation.} 
% Recent work on LLM safety has focused on preserving safety during post-training and downstream fine-tuning ~\citep{adv_finetuning, antidote}. \emph{Data- and optimization-based methods} preserve alignment through safety supervision or control of training data and optimization: SafeInstr~\citep{safeInstr} mixes safety examples into downstream fine-tuning, Lisa~\citep{lisa} alternates safety and task optimization with a proximal constraint, and SEAL~\citep{seal} selects safety-compatible fine-tuning samples through bilevel optimization. \emph{Parameter- and update-constrained methods} restrict model updates through neuron-level protection in SN-Tune and RSN-Tune ~\citep{sn-tune}, alignment-direction regularization in AsFT~\citep{asft}. \emph{Low-rank and post-hoc model-space methods} constrain or modify learned updates, including Safe LoRA~\citep{safe-lora}, SaLoRA~\citep{salora}, Safe Delta~\citep{safe-delta}, and RESTA~\citep{Resta}. However, most of these methods still attempt to protect safety within the original parameterization of the model. In contrast, our approach addresses this limitation by reparameterizing the weight space itself, enabling safety preservation through a more structured separation between safety-critical and task-adaptive components. We  discuss a broader range of methods for preserving safety during downstream adaptation in Appendix~\ref{app:additional_related_work}.
Recent efforts to mitigate safety degradation during downstream fine-tuning~\citep{adv_finetuning, antidote} broadly fall into three categories. \emph{Data- and optimization-based defenses} enforce safety via supervision or sample filtering: SafeInstr~\citep{safeInstr} mixes safety examples into downstream training, Lisa~\citep{lisa} applies proximal regularization, SEAL~\citep{seal} selects downstream data through bilevel optimization, and SPARD~\citep{chen2026spard} selects safety examples and applies corrective updates. \emph{Gradient and subspace projection methods} restrict updates to safe sub-manifolds: SafeGrad~\citep{yi2026safegrad} and SPF~\citep{zhang2026understanding} filter conflicting task gradients via alignment directions or gradient-derived subspaces, while SaLoRA~\citep{salora} and GuardSpace~\citep{zhang2026guardrail} project low-rank updates into activation-derived null spaces; SafeAnchor~\citep{guo2026safeanchor} combines Fisher-based gradient constraints with corrective replay for continual adaptation. \emph{Parameter-level protection and post-hoc methods} preserve or restore safety at the weight level: SN-Tune~\citep{sn-tune} freezes safety-critical neurons, while AsFT~\citep{asft} regularizes updates relative to an alignment direction. After fine-tuning, SafeLoRA~\citep{safe-lora} applies alignment-aware projection to LoRA updates, SafeDelta~\citep{safe-delta} selects task-delta components and compensates for residual safety loss, and RESTA~\citep{Resta} adds a safety vector through task arithmetic.
However, most of these methods still attempt to protect safety within the original parameterization of the model. In contrast, our approach addresses this limitation by reparameterizing the weight space itself, enabling safety preservation through a more structured separation between safety-critical and task-adaptive components. We  discuss a broader range of methods for preserving safety during downstream adaptation in Appendix~\ref{app:additional_related_work}.

\textbf{Analyzing safety in representation and parameter space.}
Other recent work analyzes LLM safety through structured dimensions and spaces~\citep{zou2025representationengineeringtopdownapproach, geometry_cone}. A recent study~\citep{hidden_dimensions} shows that safety-aligned behavior is not governed by a single refusal direction, but by a multi-dimensional residual space in which a dominant direction captures refusal while additional orthogonal directions encode other safety-related features. 
A complementary line of work~\citep{assessing_the_brittleness} studies safety directly in the original weight space and shows that safety-critical regions can be identified not only at the neuron level but also at the rank level.
% \textcolor{purple}{\sout{We provide a detailed discussion and comparison with this line of work in Appendix~\ref{app:actsved}.}} %\textcolor{orange}{The work uses pruning and low-rank attacks on these regions to assess the brittleness of safety alignment; we discuss how this approach differs from WSR-Tune in Appendix~\ref{app:actsved}.}
% \textcolor{purple}{At the output level, \citep{constrained_sft} find that safety alignment can alter the model's distribution mainly at the first few response tokens. And at the attention level, Surgery~\citep{liu2026surgery} associates harmful fine-tuning with changes in attention-sink behavior that distinguish harmful responses from refusals.}
Safety appears to be highly entangled with general learning rather than isolated in a distinct linear subspace, as both weight-space and activation-space analyses show substantial overlap between safety-related and utility-related directions~\citep{Safety_subspace_arenot_linearly_distinct}. Overall, these studies suggest that safety is distributed across directions, ranks, and subspaces rather than being cleanly localized to a small set of canonical parameters~\citep{geometry_cone, li2025safety}. %\jsp{위에서도 $geometry cone$ 인용되어있는데 없어도 되지 않을까 합니다.}
This in turn implies that post-hoc selection of important components in the original space may be insufficient for robust safety preservation. 

\textbf{Subspace-constrained adaptation.} Although not focused on safety, a line of prior work studies how to reduce interference by constraining parameter updates within structured subspaces. AlphaEdit \citep{alphaedit} projects updates into the null space of preserved knowledge, while orthogonal fine-tuning methods, such as PSOFT \citep{psoft} orthogonally transforms the principal subspace of pretrained weights to preserve its geometry. These methods mitigate interference by carefully selecting update directions, but they still operate within the original parameterization of the model.  In a related but distinct direction, \citep{warp} explores weight space rotation primarily in the context of class-incremental learning. While this highlights the potential of restructuring the parameter space, it is not designed to address safety preservation during downstream adaptation. These works motivate our view that the safety problem is not only about identifying certain parameters, but also about choosing a suitable space in which preservation and adaptation can be disentangled.

% \textbf{Subspace-constrained adaptation.} Although not focused on safety, prior work studies interference reduction by constraining parameter updates within structured subspaces. AlphaEdit~\citep{alphaedit} projects updates into the null space of preserved knowledge, while PSOFT~\citep{psoft} applies orthogonal transformations within a principal subspace of pretrained weights. These methods reduce interference by selecting update directions, but still operate within the model's original parameterization. 
% In a related but distinct direction, ~\citep{warp} explores weight-space rotation primarily for class-incremental learning. While this highlights the potential of restructuring the parameter space, it is not designed to address safety preservation during downstream adaptation. These works motivate our view that the safety problem is not only about identifying which parameters matter, but also about choosing a more suitable space in which preservation and adaptation can be disentangled.

\vspace{-1mm}

\section{Algorithm}
\label{sec:algorithm}

\vspace{-1mm}

\subsection{Problem setup}

\vspace{-1mm}

We consider adapting an LLM to downstream tasks while preserving its previously learned safety behaviors. Let $f_{W_0}$ denote a safety-tuned model with parameters $W_0$, where “safety-tuned” indicates that the model has learned to refuse harmful requests through prior safety post-training. In many practical settings, such a model is subsequently adapted to a downstream task using a task-specific dataset $D_{\mathrm{task}}$. However, standard fine-tuning on $D_{\mathrm{task}}$ often degrades the model’s previously learned safety behaviors, even when the downstream data itself is benign. This leads to an inherent trade-off between improving downstream performance and preserving safety during adaptation.

Formally, let $\mathcal{L}(W; D)$ denote the loss of model $W$ on dataset $D$, and let $D_{\mathrm{safe}}$ denote a safety-oriented dataset used to preserve safe behavior. In conventional fine-tuning, adaptation is performed by applying a parameter update $\Delta W$ to $W_0$ in the original weight space. Our objective is to improve downstream performance while preserving safety: 
\begin{equation} \min_{\Delta W}\ \mathcal{L}(W_0 + \Delta W; D_{\mathrm{task}}) \ \ \text{subject to} \ \  \mathcal{L}(W_0 + \Delta W; D_{\mathrm{safe}}) \leq \mathcal{L}(W_0; D_{\mathrm{safe}}) + \epsilon, 
\end{equation} 
where $\epsilon \geq 0$ denotes a tolerable increase in safety loss.

\vspace{-1mm}

\subsection{Motivation and Insight}
\vspace{-1mm}

Despite extensive efforts, existing approaches \citep{safeInstr,sn-tune,safe-lora,safe-delta,Resta} operate in the original parameter space. To understand the limitation of this representation, consider 
an arbitrary  $m\times n$ weight matrix
\begin{equation}
W =
\begin{bmatrix}
w_{11} & \cdots & w_{1n} \\
\vdots & \ddots & \vdots \\
w_{m1} & \cdots & w_{mn}
\end{bmatrix}= %\sum_{ij}w_{ij}b_{ij}
\sum_{i=1}^m\sum_{j=1}^nw_{ij}b_{ij}
\end{equation}
of a \textit{safety-tuned} model, where $\{w_{ij}\}$ are the model parameters and $\{b_{ij}\}$ are the canonical basis matrices. Each $b_{ij}$ has a single nonzero entry of 1 at position $(i,j)$ and zeros elsewhere.

In this coordinate system, safety-related information is not localized to a specific coordinate $w_{ij}$, but is instead distributed across multiple directions in the parameter space. Consequently, downstream fine-tuning (even when applied to only a subset of parameters) can still overwrite safety-related information encoded in other parameters, leading to degraded safety. This hypothesis is supported by our preliminary results in Table~\ref{tab:basis_rotation_motivation}, where  freezing a certain fraction of parameters in the original space is insufficient to adequately preserve safety. This raises the following question:

\textit{Is the original coordinate system the most suitable for fine-tuning, or can we construct a better basis that enables effective downstream  adaptation without compromising safety?}

\vspace{-1mm}
\subsection{WSR-Tune: Reparameterized Fine-Tuning for Safety Preservation}
\label{sec:wsr-algorithm}
\vspace{-1mm}
We propose WSR-Tune, which moves beyond identifying \textit{what} to protect and instead reconsiders \textit{how} the parameter space itself is represented.  To this end, we transform the weight space into a safety-conditioned coordinate system in which safety-relevant behavior is more compactly concentrated, making it easier to preserve during downstream fine-tuning. We replace the canonical parameterization 
\begin{equation}
W = \sum_{i=1}^m\sum_{j=1}^nw_{ij}b_{ij}
\;\xrightarrow{\text{Reparameterization}}\;
W = \sum_{i=1}^m\sum_{j=1}^n \tilde{w}_{ij} \phi_{ij}, 
\end{equation}
where $\{\phi_{ij}\}$ denotes a new basis and $\{\tilde{w}_{ij}\}$ are the corresponding weights in the reparameterized space. Unlike the canonical basis, where each $b_{ij}$ contains a single nonzero entry, the new basis matrices can represent structured directions spanning multiple parameters. The reparameterized weight matrix $\tilde{W}$ can be expressed in terms of the new basis $\{\phi_{ij}\}$ as
\begin{equation}
\tilde{W} =
\begin{bmatrix}
\tilde{w}_{11} & \cdots & \tilde{w}_{1n} \\
\vdots & \ddots & \vdots \\
\tilde{w}_{m1} & \cdots & \tilde{w}_{mn}
\end{bmatrix}
=
\begin{bmatrix}
\langle W, \phi_{11} \rangle & \cdots & \langle W, \phi_{1n} \rangle \\
\vdots & \ddots & \vdots \\
\langle W, \phi_{m1} \rangle & \cdots & \langle W, \phi_{mn} \rangle
\end{bmatrix},
\end{equation}
where $\langle \cdot, \cdot \rangle$ denotes the Frobenius inner product between two matrices.

This transformation aims to make safety-critical directions more identifiable and concentrated in the reparameterized space, while leaving the complementary directions with sufficient capacity for adaptation. Our key idea is to perform adaptation by updating a subset of $\tilde{w}_{ij}$, rather than directly modifying the original parameters $w_{ij}$, thereby better preserving safety during downstream adaptation. The high-level idea of WSR-Tune is illustrated in Figure~\ref{fig:WSR-Tune_figure}. In the following, we describe how our reparameterization is constructed and applied in practice.

\textbf{Safety-aware weight reparameterization.} Our first step is to construct a new basis $\{\phi_{ij}\}$ in which safety-relevant information is concentrated in a small subset of directions. Let $W_l \in \mathbb{R}^{m_l \times n_l}$ denote the weight matrix of the $l$-th selected layer. For each sample $x_i$ and token position $t \in [T_i]$, let $h_l(x_i,t) \in \mathbb{R}^{n_l}$ denote the input activation of the $t$-th token to layer $W_l$, i.e., the hidden representation of that token produced by the preceding layers before being transformed by $W_l$. We then collect these token-wise activations to form
\begin{equation}
H_l = [\, h_l(x_i,t) \,]_{(i,t)\in\mathcal{T}_{\mathrm{safe}}}
\in \mathbb{R}^{n_l \times M},
\end{equation}
where $\mathcal{T}_{\mathrm{safe}}$ denotes the set of valid sample-token indices from the safety examples in $D_{\mathrm{safe}}$, $T_i$ denotes the number of valid tokens in sample $x_i$, and $M = \sum_{i=1}^N T_i$ is the total number of collected token-wise activations. Here, $D_{\mathrm{safe}}$ consists of harmful prompts paired with safe refusal responses. We then compute the activation covariance and perform singular value decomposition:
\begin{equation}
H_l H_l^\top = U_l \Sigma_l U_l^\top,
\end{equation}

\begin{wrapfigure}{r}{0.42\textwidth}
    \vspace{-6mm}
    \centering
\includegraphics[width=0.95\linewidth]{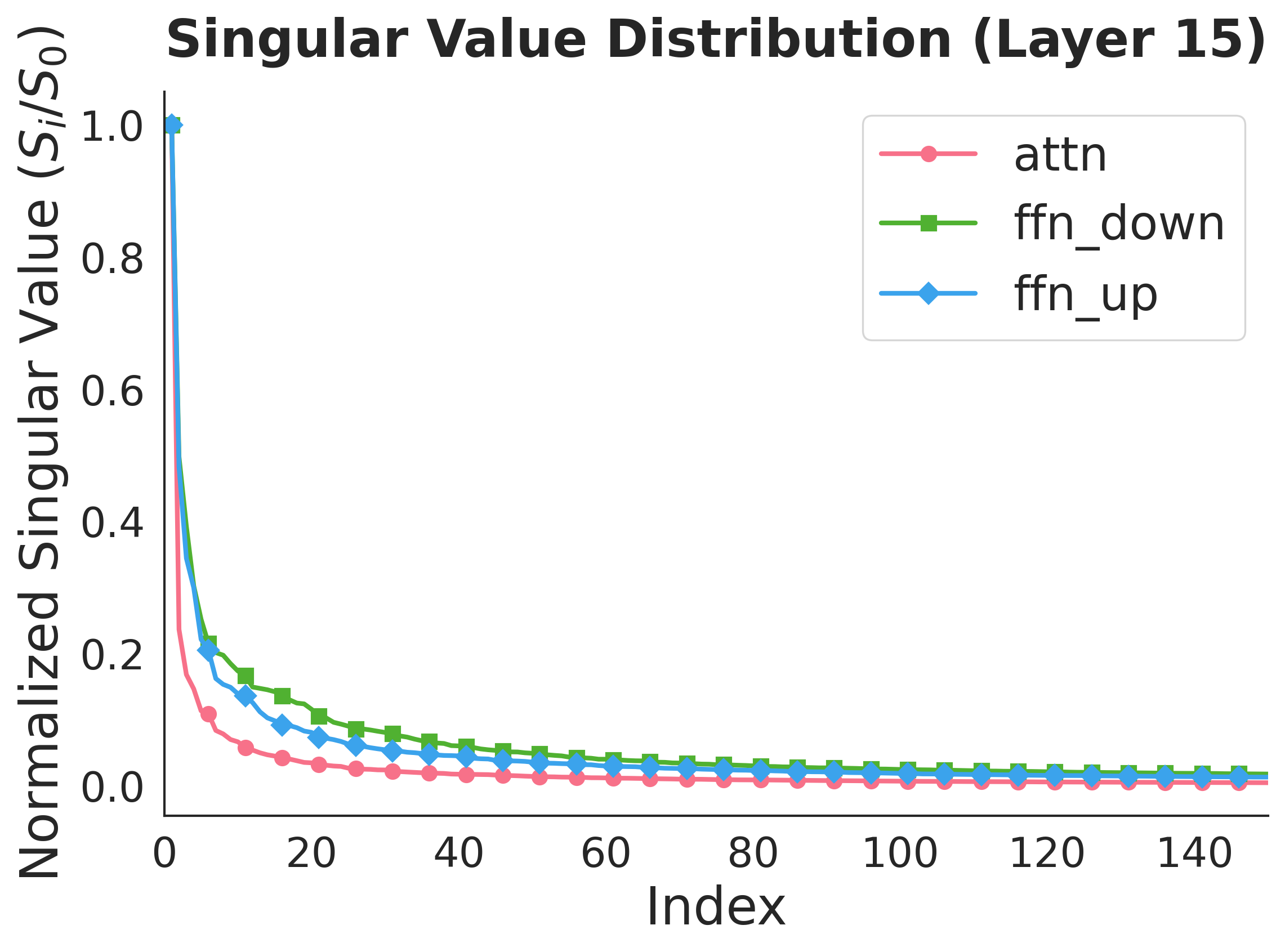}
    \caption{For each layer, singular values are sorted in descending order and normalized by the largest value. Llama-2-7B-Chat model is used. 
    }
    \label{fig:layer_15}
    \vspace{-5mm}
\end{wrapfigure}

where $U_l \in \mathbb{R}^{n_l \times n_l}$ is an orthonormal matrix whose columns are ordered according to the singular values in $\Sigma_l$. Figure~\ref{fig:layer_15} shows that the singular values exhibit a clear low-rank structure, motivating the use of $U_l$ as an activation-induced basis.

We thus reparameterize the weight matrix as 
\begin{equation} 
\label{eq:reparameterize_weight_matrix}
\widetilde{W}_l  = V_l^\top W_l U_l, 
\end{equation} 
where $U_l$ is defined above, $V_l \in \mathbb{R}^{m_l \times m_l}$ % \djh{V의 dimension 언급} 
is an arbitrary unitary matrix, and $\widetilde{W}_l \in \mathbb{R}^{m_l \times n_l}$ denotes the weights in the reparameterized space. 
Let $\{v_i\}_{i=1}^{m_l}$ and $\{u_j\}_{j=1}^{n_l}$ denote the column vectors of $V_l$ and $U_l$, respectively.  
Introducing $V_l$ allows us to define a complete orthonormal basis over the matrix space via outer products of row and column directions, i.e., 
\begin{equation}
\phi_{ij} = v_i u_j^\top, 
\end{equation}
ensuring that each $(i,j)$ corresponds to a distinct basis element. In this work, we set $V_l=I$ for simplicity,  which avoids multiple equivalent representations induced by different choices of $V_l$; Appendix~\ref{app:v_basis} shows that WSR-Tune remains robust to alternative choices of $V_l$.
%\kms{Reviewer 1의 $V_l$ matrix의 basis choice 실험} 
Under this formulation, each entry $\widetilde{W}_l(i,j)=\tilde{w}_{ij}$ represents the reparameterized weight along the basis element $\phi_{ij}$. This reparameterization defines a coordinate system aligned with activation statistics, in which safety-related information can be represented more compactly. The new coordinate framework is constructed only once before downstream fine-tuning, which is then kept fixed throughout adaptation.

\textbf{Preserving safety in the reparameterized space.} Under this reparameterization, downstream adaptation is applied to the coefficients of $\widetilde{W}_l$, rather than to $W_l$. 
To preserve safety, within each layer, we estimate the importance of each coefficient using its gradient magnitude on a safety corpus:
\begin{equation}
G_l = \sum_{x \in D_{\mathrm{safe}}} \left|\frac{\partial \mathcal{L}(x)}{\partial \widetilde{W}_l}\right|.
\end{equation}
Here, $G_l(i,j)$ quantifies the sensitivity of the safety loss to $\widetilde{W}_l(i,j)$, and larger values indicate greater importance for safety retention. Based on $G_l$, we construct a binary mask $M_l \in \{0,1\}^{m_l \times n_l}$, where $M_l(i,j)=1$ denotes a safety-critical coefficient to be preserved in the reparameterized space. %\kms{Reviewer 1이 헷갈려 한 masking이 적용되는 sapce는 in a reparameterzed space 언급.} 
In practice, we select the top-$\rho$ fraction of entries according to $G_l$ within each layer, where $\rho \in [0,1]$ controls the trade-off between safety retention and downstream adaptability. 
%\textcolor{purple}{\sout{While we use gradient magnitude for importance estimation,  WSR can accommodate alternative importance and selection mechanisms; we demonstrate this compatibility in Table~\ref{tab:baseline_plus_warp}.}}
% \wjb{"While we use gradient magnitude for importance estimation~" 부분은 제거하는게 맞는 것 같습니다. Table 3의 내용이 importance score를 다르게 설정하는 실험 내용이 아니기 때문입니다.} \kms{scoring criterion은 유지하고 importance를 측정하는 space를 변경한다는 의미로 아래처럼 수정하면 어떨까요? "WSR retains the importance scoring criterion while shifting its evaluation from the original weight space to the reparameterized space;"} \wjb{근데 지금 table 3가 Compatibility 관련 실험이라서, importance를 측정하는 space랑은 관련이 없어보여서, we demonstrate this compatibility in Table 3 부분은 삭제하는 편이 맞을 것 같습니다.} 

\textbf{Downstream adaptation in the complementary subspace.} During downstream fine-tuning, we update only the complementary coefficients while blocking gradients through the masked ones:
\begin{equation}
\widetilde{W}_l \leftarrow
\mathrm{stopgrad}(\widetilde{W}_l \odot M_l)
+
\widetilde{W}_l \odot (1 - M_l),
\end{equation}
where $\odot$ denotes element-wise multiplication. This constrains adaptation to the complementary subspace, while preserving the coefficients most critical for safety in the reparameterized space.

% \textcolor{orange}{Importantly, WSR-Tune does not introduce a new importance scoring or masking; its key distinction is the safety-conditioned coordinate system in which these operations are applied. With $V_l=I$, freezing $\widetilde W_l(i,j)$ imposes $e_i^\top\Delta W_lu_j=0$, while $\partial\mathcal L/\partial\widetilde W_l =(\partial\mathcal L/\partial W_l)U_l$. Thus, reparameterization changes both the coefficients identified as safety-critical and the admissible update geometry induced by protecting them. Appendix~\ref{app:reparam_analysis} further shows that, under the same masking rule, this coordinate change alters the induced update constraints and yields more safety-selective protection than original-space.}

%\textcolor{orange}{Our WSR-Tune differs from standard fine-tuning by restructuring the weight space for safety preservation. First, it constructs a safety-conditioned basis once and keeps it fixed throughout downstream adaptation. Second, protection is applied in this reparameterized space, where coefficients are aligned with safety-relevant activation directions. This enables targeted safety preservation while leaving the remaining directions available for downstream updates. Appendix~\ref{app:reparam_analysis} further examines how this coordinate change enables more safety-selective protection under the same masking rule.}

Our WSR-Tune differs from standard fine-tuning by reparameterizing the weight space for safety preservation. First, it decouples basis construction from task adaptation: a safety-conditioned basis is constructed once using safety data and remains fixed throughout downstream adaptation. Second, protection is applied in the reparameterized space, where coefficients are aligned with safety-relevant activation directions. This enables targeted safety preservation while leaving the remaining directions available for downstream updates, thereby reducing interference between safety preservation and downstream adaptation. Appendix~\ref{app:reparam_analysis} further examines how this coordinate change enables more safety-selective protection under the same masking rule.

% Our WSR-Tune differs from standard fine-tuning in two key aspects. First, it decouples basis construction from task adaptation: a safety-conditioned basis is constructed once using safety data and kept fixed thereafter. Second, protection is applied in the reparameterized space, where coefficients are aligned with safety-relevant activation directions. This enables more targeted preservation of safety and reduces interference during downstream fine-tuning. 
%\kms{Reviewer 3의 W1/Q1에서 작성한 novelty는 reparameterization이 같은 protection rule의 importance ranking과 admissible-update geometry를 바꾼다는 것을 강조}

% \wjb{별도의 subsection으로 WSR-LoRA를 빼기보다, 3-3의 paragraph로 빼서 PEFT 적용이 가능함을 간단하게 서술하면 어떨까 생각함. 현재 수식의 경우 GuardSpace 및 PiSSA 방식과 너무 일치해보인다는 인식을 줄 것 같다는 생각임.}

\subsection{Discussion}

%We discuss several practical aspects of WSR-Tune.
%\wjb{위 한 줄은 없어도 될 것 같습니다.}

%\jsp{Discussion의 의미가 애매한 것 같아서 아예 빼는 건 어떨까 싶습니다. Applicability는 implementation details로 옮기고 compatibility, freeze ratio는 해당 본문에서 설명하면 되지 않나라는 생각입니다.}

\textbf{Extension to parameter-efficient fine-tuning.}
\label{sec:wsr_lora}
% 기존 문단: We extend WSR-Tune to parameter-efficient fine-tuning (PEFT) by applying safety-conditioned reparameterization and protection to LoRA~\citep{lora} factors. Detailed implementation and optimization procedures are provided in Appendix~\ref{app:wsr_lora}.
To support practical downstream adaptation under limited memory and compute budgets, we extend WSR-Tune beyond full-parameter fine-tuning to parameter-efficient fine-tuning. We apply safety-conditioned reparameterization and protection to LoRA~\citep{lora} factors, enabling safety-preserving adaptation with a small set of trainable parameters. Detailed implementation and optimization procedures are provided in Appendix~\ref{app:wsr_lora}.
% \jsp{lora와의 결합을 보여주는 이유가 있으면 좋을 것 같습니다. Due to the practical usage of peft methods... 또는 We show that our method is compatible to lora ... 또는 to demonstrate that the proposed reparameterization principle is not restricted to full-parameter adaptation ...} \wjb{좋은 것 같습니다.} \kms{반영하여 업데이트했습니다.}

\textbf{Applicability to different layers.} WSR-Tune can be applied not only to MLP layers but also to attention layers, including the key, query, and value projection matrices. In our experiments in Section \ref{sec:experiments}, we apply WSR-Tune to both attention and MLP projections and evaluate its effectiveness. % \wjb{당연한 이야기라서, 이 paragraph의 내용이 필요할까 싶은 생각입니다. 만약 모든 decoder layer에 적용되기 어려운 방법론이 있었다면 의미가 있겠지만, 당연한 내용이라 굳이 서술할 필요는 전혀 없어보입니다. 굳이 서술한다면, "not only to attention layers but also to MLP layers"라는 표현이 더 좋다는 생각입니다. (하지만 그럼에도 이 paragraph 를 서술할 이유가 없다고 생각은 합니다.} \kms{특정 module에 특화되는 방법론이 아니고, appendix A에서 wsr-tune에 적용하는 layer module에 대해서 언급하였기 때문에 저도 이 paragraph의 내용은 굳이 서술하지 않아도 된다 생각합니다.}

\textbf{Compatibility with prior works.}
% WSR-Tune can be integrated with existing methods such as SafeInstr, SN-Tune, and SEAL by applying their native protection mechanisms within the reparameterized weight space. We validate this in Section~\ref{sec:experiments}.
We validate this in Section~\ref{sec:experiments}, which highlights the plug-and-play compatibility of WSR-Tune, allowing it to seamlessly enhance prior safety-preserving methods without altering their native mechanisms.
% \wjb{Prior work와의 plug-and-play가 가능하다는 표현을 추가하면 어떨까 싶습니다: We validate this in Section~\ref{sec:experiments}, which highlights the plug-and-play compatibility of WSR-Tune, allowing it to seamlessly enhance prior safety-preserving methods without altering their native mechanisms. 이 표현을 사용한다면, 굳이 such as ~ 로 method 명을 기입하지 않아도 될 것 같습니다. (특정 method에 한정해서만 compatibility가 성립한다는 느낌을 준다는 생각입니다.}

\textbf{Freeze ratio.} The freeze ratio $\rho$ controls the fraction of coefficients preserved for safety. In Section~\ref{sec:experiments}, we study the impact of varying $\rho$, demonstrating its role in balancing safety and downstream performance. Notably, WSR-Tune achieves a strong trade-off even with relatively small values of $\rho$.

%\textcolor{orange}{\textbf{Computational overhead.} WSR-Tune constructs the safety-conditioned basis via SVD once and reuses it throughout fine-tuning. In our runtime evaluation, this one-time step takes only about 24\% of the three-epoch downstream fine-tuning time (Appendix~\ref{app:runtime}).}

\textbf{Computational overhead.} WSR-Tune requires a one-time SVD to construct the safety-conditioned basis, which is then fixed and reused throughout downstream fine-tuning. Thus, the basis construction introduces only a one-time preprocessing cost, without requiring repeated decomposition or additional optimization during adaptation. We provide a detailed runtime analysis in Appendix~\ref{app:runtime}.

\begin{table*}[t]
\vspace{-4mm}
\centering
\scriptsize
\caption{Comparison across various chat and instruction-tuned models. Safety is measured by Jailbreak (JB) AVG (averaged over Direct, AutoDAN, PAIR, and PAP; lower is better), and downstream performance is measured by task accuracy (higher is better). Specifically, we use GSM8K for the Llama-2 Chat models and MATH for the Llama-3 Instruct models.
We also report changes relative to Full Params FT: $\Delta_{\mathrm{S}}$ (safety, lower is better) and $\Delta_{\mathrm{D}}$ (downstream, higher is better). We further define $\Delta_{\mathrm{Overall}} = -\Delta_{\mathrm{S}} + \Delta_{\mathrm{D}}$ to summarize the safety--utility trade-off.
}
\label{tab:main_all_models_delta}
\resizebox{\textwidth}{!}{
\begin{tabular}{lccccc|ccccc}
\toprule
& \multicolumn{5}{c|}{\textbf{Llama-2-7B-Chat}} 
& \multicolumn{5}{c}{\textbf{Llama-2-13B-Chat}} \\
\cmidrule(lr){2-6} \cmidrule(lr){7-11}
& \multicolumn{2}{c}{\textbf{Safety $\downarrow$}}
& \multicolumn{2}{c}{\textbf{Downstream $\uparrow$}}
& \multicolumn{1}{c|}{\textbf{Overall} $\uparrow$}
& \multicolumn{2}{c}{\textbf{Safety $\downarrow$}}
& \multicolumn{2}{c}{\textbf{Downstream $\uparrow$}}
& \multicolumn{1}{c}{\textbf{Overall} $\uparrow$} \\
\cmidrule(lr){2-3} \cmidrule(lr){4-5} \cmidrule(lr){6-6}
\cmidrule(lr){7-8} \cmidrule(lr){9-10} \cmidrule(lr){11-11}
\textbf{Method}
& \textbf{JB AVG} & $\boldsymbol{\Delta_{\mathrm{S}}}$
& \textbf{Accuracy} & $\boldsymbol{\Delta_{\mathrm{D}}}$ & $\boldsymbol{\Delta_{\mathrm{Overall}}}$
& \textbf{JB AVG} & $\boldsymbol{\Delta_{\mathrm{S}}}$
& \textbf{Accuracy} & $\boldsymbol{\Delta_{\mathrm{D}}}$ & $\boldsymbol{\Delta_{\mathrm{Overall}}}$ \\
\midrule

Full Params FT
& 20.78 & 0.00 & 41.17 & 0.00 & 0.00
& 10.04 & 0.00 & 45.94 & 0.00 & 0.00 \\
\midrule

SafeInstr
& 13.55 & -7.23 & 38.44 & -2.73 & +4.50
& 5.82 & -4.22 & 46.55 & +0.61 &  +4.83  \\

Resta
& 10.56 & -10.22 & 36.92 & -4.25 &  +5.97 
& 10.34 & +0.30 & 44.20 & -1.74 & -2.04 \\

SEAL
& 12.75 & -8.03 & 38.89 & -2.28 & +5.75
& 10.28 & +0.24 & 45.19 & -0.75 & -0.99 \\

AsFT
& 5.27 & -15.51 & 19.71 & -21.46 & -5.95
& 0.86 & -9.18 & 28.43 & -17.51 & -8.33 \\

Lisa
& 9.95 & -10.83 & 14.40 & -26.77 & -15.94
& 1.39 & -8.65 & 36.24 & -9.70 & -1.05 \\

SafeDelta
& 5.57 &  -15.21  & 19.79 & -21.38 & -6.17
& 0.97 &  -9.07  & 28.43 & -17.51 & -8.44 \\

SN-Tune
& 27.18 & +6.40 & 38.99 &  -2.18  & -8.58
& 28.97 & +18.93 & 48.22 & +2.28 & -16.65 \\

RSN-Tune
& 20.73 & -0.05 & 40.26 & -0.91 & -0.86
& 28.79 & +18.75 & 49.96 &  +4.02 & -14.73 \\

\midrule

\textbf{WSR-Tune (Ours)}
& 6.90 & -13.88 & 38.99 &  -2.18 & \cellcolor{green!20}\textbf{+11.70}
& 1.35 &  -8.69  & 49.58 &  +3.64  & \cellcolor{green!20} \textbf{+12.33} \\

\bottomrule
\end{tabular}
}

\vspace{1.0em}

\resizebox{\textwidth}{!}{
\begin{tabular}{lccccc|ccccc}
\toprule
& \multicolumn{5}{c|}{\textbf{Llama-3.2-3B-Instruct}} 
& \multicolumn{5}{c}{\textbf{Llama-3.1-8B-Instruct}} \\
\cmidrule(lr){2-6} \cmidrule(lr){7-11}
& \multicolumn{2}{c}{\textbf{Safety $\downarrow$}}
& \multicolumn{2}{c}{\textbf{Downstream $\uparrow$}}
& \multicolumn{1}{c|}{\textbf{Overall} $\uparrow$}
& \multicolumn{2}{c}{\textbf{Safety $\downarrow$}}
& \multicolumn{2}{c}{\textbf{Downstream $\uparrow$}}
& \multicolumn{1}{c}{\textbf{Overall} $\uparrow$} \\
\cmidrule(lr){2-3} \cmidrule(lr){4-5} \cmidrule(lr){6-6}
\cmidrule(lr){7-8} \cmidrule(lr){9-10} \cmidrule(lr){11-11}
\textbf{Method}
& \textbf{JB AVG} & $\boldsymbol{\Delta_{\mathrm{S}}}$
& \textbf{Accuracy} & $\boldsymbol{\Delta_{\mathrm{D}}}$ & $\boldsymbol{\Delta_{\mathrm{Overall}}}$
& \textbf{JB AVG} & $\boldsymbol{\Delta_{\mathrm{S}}}$
& \textbf{Accuracy} & $\boldsymbol{\Delta_{\mathrm{D}}}$ & $\boldsymbol{\Delta_{\mathrm{Overall}}}$ \\
\midrule

Full Params FT
& 8.65 & 0.00 & 21.52 & 0.00 & 0.00
& 9.28 & 0.00 & 12.12 & 0.00 & 0.00 \\
\midrule

SafeInstr
& 8.23 & -0.42 & 22.26 & +0.74 &  +1.16
& 6.21 & -3.07 & 13.98 & +1.86 &  +4.93 \\

Resta
& 8.12 & -0.53 & 20.56 & -0.96 & -0.43
& 6.74 & -2.54 & 10.56 & -1.56 & +0.98 \\

SEAL
& 8.37 & -0.28 & 21.62 & +0.10 & +0.38
& 5.45 & -3.83 & 13.18 & +1.06 & +4.89 \\

AsFT
& 7.11 & -1.54 & 7.76 & -13.76 & -12.22
& 3.36 & -5.92 & 5.66 & -6.46 & -0.54 \\

Lisa
& 6.83 & -1.82 & 8.82 & -12.70 & -10.88
& 3.52 & -5.76 & 11.32 & -0.80 & +4.96 \\

SafeDelta
& 6.48 &  -2.17 & 6.48 & -15.04 & -12.87
& 4.55 &  -4.73 & 1.18 & -10.94 & -6.21 \\

SN-Tune
& 21.84 & +13.19 & 19.36 & -2.16 & -15.35
& 54.35 & +45.07 & 13.62 & +1.50 & -43.57 \\

RSN-Tune
& 24.49 & +15.84 & 19.62 & -1.90 & -17.74
& 55.34 & +46.06 & 14.10 & +1.98 & -44.08 \\

\midrule

\textbf{WSR-Tune (Ours)}
& 6.40 &  -2.25  & 22.38 &  +0.86  & \cellcolor{green!20}\textbf{+3.11}
& 5.62 &  -3.66 & 13.70 & +1.58 & \cellcolor{green!20}\textbf{+5.24} \\

\bottomrule
\end{tabular}
}
%\vspace{-5mm}
\end{table*}

%\vspace{-2mm}
\section{Experiments}
\label{sec:experiments}
%\vspace{-2mm}

%\wjb{main table 실험의 경우 AsFT와 Lisa는 WSR-Tune의 baseline으로 올려도 되지 않을까 생각듭니다. LoRA setting이기는 했지만, WSR-Tune보다 열등하기 때문에 괜찮지 않을까 싶습니다.}
% \kms{\wjb: downstream performance보다 task performance가 더 나아보입니다. 정확히는 downstream task performance인데 reviewer들이 task performance를 더 사용했습니다.}

\subsection{Experimental Setup \& Details}
\vspace{-1mm}

\textbf{Datasets and models.} Unless otherwise specified, we use the Circuit Breakers dataset~\citep{circuit_breakers} as the safety dataset, consisting of harmful prompts with safe refusal responses. To evaluate sensitivity to the safety-data source, we conduct additional experiments using safety data selected from BeaverTails~\citep{beavertails} in Appendix~\ref{app:safety_data}. For downstream adaptation, we use GSM8K~\citep{gsm8k}, Hendrycks MATH~\citep{hendrycksmath}, MedQA~\citep{medqa}, and ARC-Challenge~\citep{arc-c}. We additionally evaluate AGNEWS~\citep{agnews} and report the results in Appendix~\ref{app:agnews_results}, extending the downstream evaluation beyond reasoning tasks. 
We conduct experiments on a diverse set of large language models spanning the Llama~\citep{llama2, llama3}, Qwen~\citep{qwen2.5}, and Gemma~\citep{gemma2} families, where the exact configurations are provided in  Appendix~\ref{app:experimental_setup}. %기존 문장 and report the exact model configurations in the corresponding result tables.
%\textcolor{purple}{and exact configurations are provided in Appendix~\ref{app:experimental_setup}.}

%\textbf{Baselines.} In the full-parameter setting, we compare WSR-Tune with Full Params FT, SafeInstr~\citep{safeInstr}, Resta~\citep{Resta}, SafeDelta~\citep{safe-delta}, SEAL~\citep{seal}, SN-Tune, and RSN-Tune~\citep{sn-tune}. Full Params FT updates all weights without safety constraints. SafeInstr augments downstream data with safety examples, SEAL selects safety-compatible training samples, Resta and SafeDelta modify fine-tuned weights post hoc, and SN-Tune and RSN-Tune constrain updates using safety-neuron importance. For PEFT, we compare WSR-LoRA with Vanilla LoRA FT, AsFT~\citep{asft}, Lisa~\citep{lisa}, SafeLoRA~\citep{safe-lora}, and SaLoRA~\citep{salora}. Vanilla LoRA FT trains low-rank adapters without safety constraints; AsFT regularizes updates using alignment directions; Lisa alternates safety and task optimization with a proximal penalty; SafeLoRA projects learned updates after training; and SaLoRA combines a fixed safety-preserving projection with task-specific initialization. AsFT and Lisa are evaluated with LoRA adapters in our experiments.

% \kms{Full FT, PEFT 별로 각각의 메소드에 대해 간단하게 설명}

\textbf{Baselines.} We compare WSR-Tune with representative safety-preserving fine-tuning baselines; %full-parameter setting \djh{full-parameter setting 삭제??}, 
SafeInstr~\citep{safeInstr} and SEAL~\citep{seal}, which use safety supervision or select safety-compatible data; Resta~\citep{Resta} and SafeDelta~\citep{safe-delta}, which modify fine-tuned weights post hoc; and SN-Tune and RSN-Tune~\citep{sn-tune}, which constrain updates using safety-neuron importance. We also consider AsFT~\citep{asft}, which regularizes updates using alignment directions, and Lisa~\citep{lisa}, %\djh{Lisa citation 확인 필요. 다시 생각해보니 Lisa, AsFT를 LoRA로 implement했다는 이야기는 initial submission에서는 굳이 안해도 될것 같습니다.}
which alternates safety and task optimization with a proximal penalty. 
%We implement AsFT and Lisa with LoRA, following their original implementations.}
% \textcolor{orange}{For PEFT, we compare WSR-LoRA with unconstrained Vanilla LoRA fine-tuning; AsFT~\citep{asft}, which regularizes updates using alignment directions; Lisa~\citep{lisa}, which alternates safety and task optimization with a proximal penalty; SafeLoRA~\citep{safe-lora}, which projects learned updates after training; and SaLoRA~\citep{salora}, which combines a fixed safety-preserving projection with task-specific initialization. AsFT and Lisa are evaluated with LoRA adapters.}

% \begin{wrapfigure}{r}{0.50\textwidth}
%     \vspace{-8mm}
%     \centering
%     \includegraphics[width=\linewidth]{figures/llama_2_13b_chat_gsm8k_tradeoff.pdf}
%     \vspace{-5mm}
%     \caption{Safety-downstream trade-off among fine-tuning on Llama-2-13B-Chat. The x-axis shows downstream accuracy, and the y-axis shows defense success rate ($1 -$ ASR), where higher indicates better safety.}
%     \label{fig:main_all_figure}
%     \vspace{-3mm}
% \end{wrapfigure}

\textbf{Evaluation metrics.}
We evaluate downstream performance using 5-shot accuracy on each downstream dataset. For safety evaluation, we use harmful prompts from AdvBench~\citep{advbench} and report attack success rate (ASR). We consider one direct prompting setting using the original AdvBench prompts, and three jailbreak attack methods, AutoDAN~\citep{autodan}, PAIR~\citep{pair}, and PAP~\citep{pap}, which generate attack prompts. We generate jailbreak prompts using the HarmBench~\citep{harmbench} and measure ASR with its refusal keywords, counting an attack as successful if the response does not begin with these keywords. We additionally evaluate the same responses using WildGuard~\citep{wildguard}, an LLM-based judge, in Appendix~\ref{sec:wildguard_judge}.
% \djh{다른 방법으로 measure한 결과들도 appendix에 있다고 언급}

% \kms{실험의 하이퍼 파라미터와 method 별 하이퍼 파라미터는 Appendix A에 작성}

\textbf{Implementation details.}
Unless otherwise specified, all methods follow the same downstream fine-tuning pipeline and are trained for 3 epochs, with method-specific learning rates chosen to balance downstream performance and safety retention. %WSR-Tune and WSR-LoRA use the same safety-data source for basis construction and importance estimation. 
Shared training settings and method-specific hyperparameters are provided in Appendix~\ref{app:hyperparameters}.

% \begin{wrapfigure}{r}{0.47\textwidth}
%     % \vspace{-8mm}
%     \centering
%     \includegraphics[width=\linewidth]{figures/llama_2_13b_chat_gsm8k_tradeoff.pdf}
%         \vspace{-5mm}
%     \caption{Safety-downstream trade-off among fine-tuning on Llama-2-13B-Chat. The x-axis shows downstream accuracy, and the y-axis shows defense success rate ($1 -$ ASR), where higher indicates better safety.}
%     \label{fig:main_all_figure}
%     % \vspace{-3mm}
% \end{wrapfigure}

%full hyperparameters are provided in the Appendix~\ref{app:hyperparameters}. SafeInstr updates all model weights using the augmented training set, whereas Resta and SafeDelta are applied post hoc to the downstream-tuned checkpoint. For SN-Tune, RSN-Tune, and WSR-Tune, we target all transformer layers and apply the method to the attention projections \texttt{q}, \texttt{k}, and \texttt{v} and the MLP projections \texttt{up} and \texttt{down}. WSR-Tune constructs the safety-conditioned basis from the Circuit Breakers dataset and uses a freeze ratio of 10\% in the reparameterized space. The effect of varying the freeze ratio is further analyzed.

% \kms{각 method 별 하이퍼 파라미터 및 세부 내용은 appendix로 보냈습니다.}

% \wjb{llama 3b table의 경우 appendix로 빼고, results 부분에서 설명할때만 서술하면 어떨까 싶음. 테이블 위치만 알려주면 됨.}

\vspace{-1mm}
\subsection{Main Results}
%\vspace{-1mm}

\begin{wrapfigure}{r}{0.50\textwidth}
    \vspace{-8mm}
    \centering
    \includegraphics[width=\linewidth]{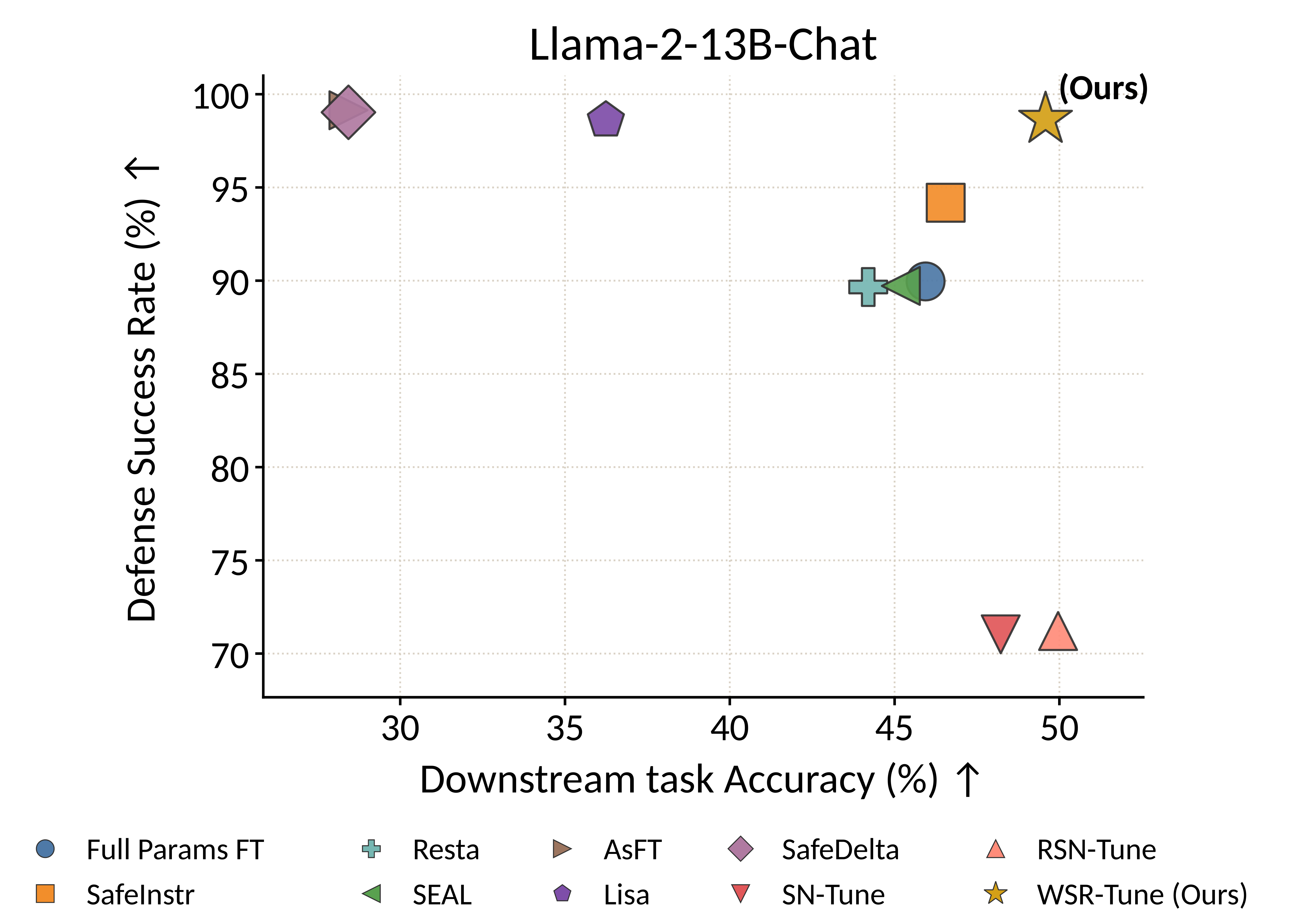}
    \vspace{-5mm}
    \caption{Safety-downstream trade-off among fine-tuning on Llama-2-13B-Chat. The x-axis shows downstream accuracy, and the y-axis shows defense success rate ($1 -$ ASR), where higher indicates better safety.}
    \label{fig:main_all_figure}
    \vspace{-3mm}
\end{wrapfigure}

\textbf{Comparison with baselines.}
% We first compare WSR-Tune with representative safety-preserving fine-tuning baselines under downstream adaptation. Table~\ref{tab:main_all_models_delta} reports the main results across chat and instruction-tuned models. Overall, WSR-Tune achieves the best trade-off between safety and downstream performance across all model settings. WSR-Tune substantially reduces jailbreak attack success while maintaining competitive or improved downstream performance, leading to the largest overall gains over full-parameter fine-tuning.  
%  Figure~\ref{fig:main_all_figure} further visualizes this trade-off. Compared with existing baselines, WSR-Tune is positioned closer to the upper-right region, indicating higher downstream accuracy and stronger defense success rate. This further shows that WSR-Tune achieves a favorable balance between safety retention and downstream adaptation. These results indicate that reparameterizing the weight space enables more effective safety preservation than directly constraining updates in the original parameter space.
We first compare WSR-Tune with representative safety-preserving fine-tuning baselines under downstream adaptation. Table~\ref{tab:main_all_models_delta} reports the main results across chat % and instruction-tuned 
models. Overall, WSR-Tune achieves the best trade-off between safety and downstream performance across all model settings. WSR-Tune substantially reduces jailbreak attack success while maintaining competitive or improved downstream performance, leading to the largest overall gains over full-parameter fine-tuning.
Figure~\ref{fig:main_all_figure} further visualizes these trade-off. Compared with existing baselines, WSR-Tune is positioned closer to the upper-right region, indicating higher downstream accuracy and stronger defense success rate, achieving a favorable balance between safety retention and downstream adaptation. These results indicate that reparameterizing the weight space enables more effective safety preservation than directly constraining updates in the original parameter space. 
% Results on Llama-3.2-3B-Instruct and Llama-3.1-8B-Instruct under MATH fine-tuning are provided in Appendix~\ref{app:llama3_results}. 
WSR-Tune also improves the safety-utility trade-off over their respective unconstrained fine-tuning baselines on both models.

% \textbf{Results on base models.} 
% We also evaluate whether the same trend holds when downstream adaptation starts from base models rather than chat or instruction-tuned models. The detailed results are provided in the Appendix~\ref{app:base_model_results}, where WSR-Tune consistently achieves the best performance. 
% This suggests that the advantage of weight-space reparameterization is not limited to a specific alignment regime, but also extends to base-to-safe-to-downstream adaptation pipelines.

% \begin{wrapfigure}{r}{0.47\textwidth}
%     \vspace{-8mm}
%     \centering
%     \includegraphics[width=\linewidth]{figures/llama_2_13b_chat_gsm8k_tradeoff.pdf}
%         \vspace{-5mm}
%     \caption{Safety-downstream trade-off among fine-tuning on Llama-2-13B-Chat. The x-axis shows downstream accuracy, and the y-axis shows defense success rate ($1 -$ ASR), where higher indicates better safety.}
%     \label{fig:main_all_figure}
%     \vspace{-3mm}
% \end{wrapfigure}

\textbf{Compatibility with existing fine-tuning methods.}
A useful property of WSR-Tune is that it can be combined with existing safety-preserving fine-tuning strategies rather than replacing them entirely. Table~\ref{tab:baseline_plus_warp} shows the performance changes after applying WSR-Tune on top of existing baselines. Across SN-Tune, SafeInstr, and SEAL, adding WSR-Tune reduces the average attack success rate while also improving downstream performance.  These results suggest that the benefits of reparameterization are complementary to existing parameter-selection and safety-data-based strategies, allowing WSR-Tune to serve as a plug-in mechanism for improving the safety-downstream trade-off of prior methods.

\begin{table*}[t]
\vspace{-6mm}
\centering
\tiny
\caption{Compatibility of WSR-Tune with existing safety-preserving fine-tuning methods on Llama2-Chat models. Lower is better for safety metrics, and higher is better for downstream performance. Values are reported as percentages without the percent sign.}
\label{tab:baseline_plus_warp}

\resizebox{0.88\textwidth}{!}{
\begin{tabular}{lcccccc}
\toprule
\textbf{Llama-2-7B-Chat} & \multicolumn{5}{c}{\textbf{Safety $\downarrow$}} & \textbf{Downstream $\uparrow$} \\
\cmidrule(lr){2-6}\cmidrule(lr){7-7}
\textbf{Method} & \textbf{Direct} & \textbf{AutoDAN} & \textbf{PAIR} & \textbf{PAP} & \textbf{AVG} & \textbf{Accuracy} \\
\midrule

SN-Tune & 0.00 & 17.69 & 45.58 & 45.46 & 27.18 & 38.99 \\
%\rowcolor{green!12} 
\rowcolor{green!12}SN-Tune + WSR-Tune & \textbf{0.00} & \textbf{11.92} & \textbf{41.92} & \textbf{38.46} & \textbf{23.07} & \textbf{41.02} \\

\midrule

SafeInstr & 0.00 & 0.38 & 18.85 & 34.96 & 13.55 & 38.44 \\
%\rowcolor{green!12} 
\rowcolor{green!12}SafeInstr + WSR-Tune & \textbf{0.00} & \textbf{0.00} & \textbf{8.46} & \textbf{29.69} & \textbf{9.53} & \textbf{39.27} \\

\midrule

SEAL
& 0.00 & 4.23 & 16.92 & 29.85 & 12.75 & 38.89 \\
\rowcolor{green!12}
SEAL + WSR-Tune
& \textbf{0.00} & \textbf{0.00} & \textbf{15.38} & \textbf{23.46}
& \textbf{9.71} & \textbf{39.13} \\

\bottomrule
\end{tabular}
}

\vspace{1.0em}

\resizebox{0.88\textwidth}{!}{
\begin{tabular}{lcccccc}
\toprule
\textbf{Llama-2-13B-Chat} & \multicolumn{5}{c}{\textbf{Safety $\downarrow$}} & \textbf{Downstream $\uparrow$} \\
\cmidrule(lr){2-6}\cmidrule(lr){7-7}
\textbf{Method} & \textbf{Direct} & \textbf{AutoDAN} & \textbf{PAIR} & \textbf{PAP} & \textbf{AVG} & \textbf{Accuracy} \\
\midrule

SN-Tune & \textbf{0} & 24.42 & \textbf{33.08} & \textbf{58.38} & 28.97 & 48.22 \\
\rowcolor{green!12} SN-Tune + WSR-Tune & \textbf{0} & \textbf{7.12} & 33.27 & \textbf{58.38} & \textbf{24.69} & \textbf{48.75} \\

\midrule

SafeInstr & \textbf{0} & \textbf{0} & 9.62 & 13.65 & 5.82 & 46.55 \\
\rowcolor{green!12} SafeInstr + WSR-Tune & \textbf{0} & \textbf{0} & \textbf{6.54} & \textbf{5.19} & \textbf{2.93} & \textbf{48.60} \\

\midrule

SEAL
& \textbf{0} & 0.96 & 15.19 & 24.96 & 10.28 & 45.19 \\
\rowcolor{green!12}
SEAL + WSR-Tune
& \textbf{0} & \textbf{0} & \textbf{2.69} & \textbf{1.31}
& \textbf{1.00} & \textbf{48.67} \\

\bottomrule
\end{tabular}
}
\vspace{-5mm}
\end{table*}

\begin{table*}[t]
\centering
\scriptsize
\caption{Additional results on Qwen-2.5-7B-Instruct and Gemma-2-9B-IT. Both models are fine-tuned on GSM8K as the downstream task.  The table follows the same delta-based format as Table~\ref{tab:main_all_models_delta}, all deltas are relative to Full Params FT.}
\label{tab:other_two_instruct_models_delta}
\resizebox{\textwidth}{!}{
\begin{tabular}{lccccc|ccccc}
\toprule
& \multicolumn{5}{c|}{\textbf{Qwen-2.5-7B-Instruct}} 
& \multicolumn{5}{c}{\textbf{Gemma-2-9B-IT}} \\
\cmidrule(lr){2-6} \cmidrule(lr){7-11}
& \multicolumn{2}{c}{\textbf{Safety $\downarrow$}}
& \multicolumn{2}{c}{\textbf{Downstream $\uparrow$}}
& \multicolumn{1}{c|}{\textbf{Overall}}
& \multicolumn{2}{c}{\textbf{Safety $\downarrow$}}
& \multicolumn{2}{c}{\textbf{Downstream $\uparrow$}}
& \multicolumn{1}{c}{\textbf{Overall}} \\
\cmidrule(lr){2-3} \cmidrule(lr){4-5} \cmidrule(lr){6-6}
\cmidrule(lr){7-8} \cmidrule(lr){9-10} \cmidrule(lr){11-11}
\textbf{Method}
& \textbf{JB AVG} & $\boldsymbol{\Delta_{\mathrm{S}}}$ 
& \textbf{Accuracy} & $\boldsymbol{\Delta_{\mathrm{D}}}$ & $\boldsymbol{\Delta_{\mathrm{Overall}}}$ 
& \textbf{JB AVG} & $\boldsymbol{\Delta_{\mathrm{S}}}$ 
& \textbf{Accuracy} & $\boldsymbol{\Delta_{\mathrm{D}}}$ & $\boldsymbol{\Delta_{\mathrm{Overall}}}$ \\
\midrule

Full Params FT
& 3.62 & 0.00 & 67.32 & 0.00 & 0.00
& 5.37 & 0.00 & 69.75 & 0.00 & 0.00 \\
\midrule

SafeInstr
& 6.66 & +3.04 & 67.70 & +0.38 & -2.66
& 8.47 & +3.10 & 70.89 & +1.14 & -1.96 \\

Resta
& 1.82 & -1.80 & 66.87 & -0.45 & +1.35
& 4.58 & -0.79 & 57.85 & -11.90 & -11.11 \\

SEAL
& 4.34 & +0.72 & 70.05 & +2.73 & +2.01
& 5.19 & -0.18 & 68.31 & -1.44 & -1.26 \\

AsFT
& 3.72 & +0.10 & 71.57 & +4.25 & +4.15
& 4.77 & -0.60 & 36.62 & -33.13 & -32.53 \\

Lisa
& 4.16 & +0.54 & 70.89 & +3.57 & +3.03
& 3.03 & -2.34 & 65.28 & -4.47 & -2.13 \\

SafeDelta
& 3.82 & +0.20 & 66.87 & -0.45 & -0.65
& 4.52 & -0.85 & 64.37 & -5.38 & -4.53 \\

SN-Tune
& 37.55 & +33.93 & 69.52 & +2.20 & -31.73
& 36.87 & +31.50 & 74.37 & +4.62 & -26.88 \\

RSN-Tune
& 38.31 & +34.69 & 71.34 & +4.02 & -30.67
& 41.15 & +35.78 & 71.57 & +1.82 & -33.96 \\
%\midrule
\midrule
\textbf{WSR-Tune (Ours)}
& 2.68 & -0.94 & 70.81 & +3.49 & \cellcolor{green!20}\textbf{+4.43}
& 4.99 & -0.38 & 70.81 & +1.06 & \cellcolor{green!20}\textbf{+1.44} \\

\bottomrule
\end{tabular}
}
\vspace{-4mm}
\end{table*}

\subsection{Additional Analyses}
\vspace{-1mm}

\textbf{Applications to other models.}  %\djh{Table 4 간단하게만 설명}
We further evaluate WSR-Tune on Qwen-2.5-7B-Instruct and Gemma-2-9B-IT, as shown in Table~\ref{tab:other_two_instruct_models_delta}. Both methods preserve safety while maintaining strong downstream performance, suggesting that the benefits of weight-space reparameterization generalize across diverse instruction-tuned model families beyond Llama architectures.

\begin{figure}[t]
\vspace{2.5mm}
    \centering
    \includegraphics[width=\textwidth]{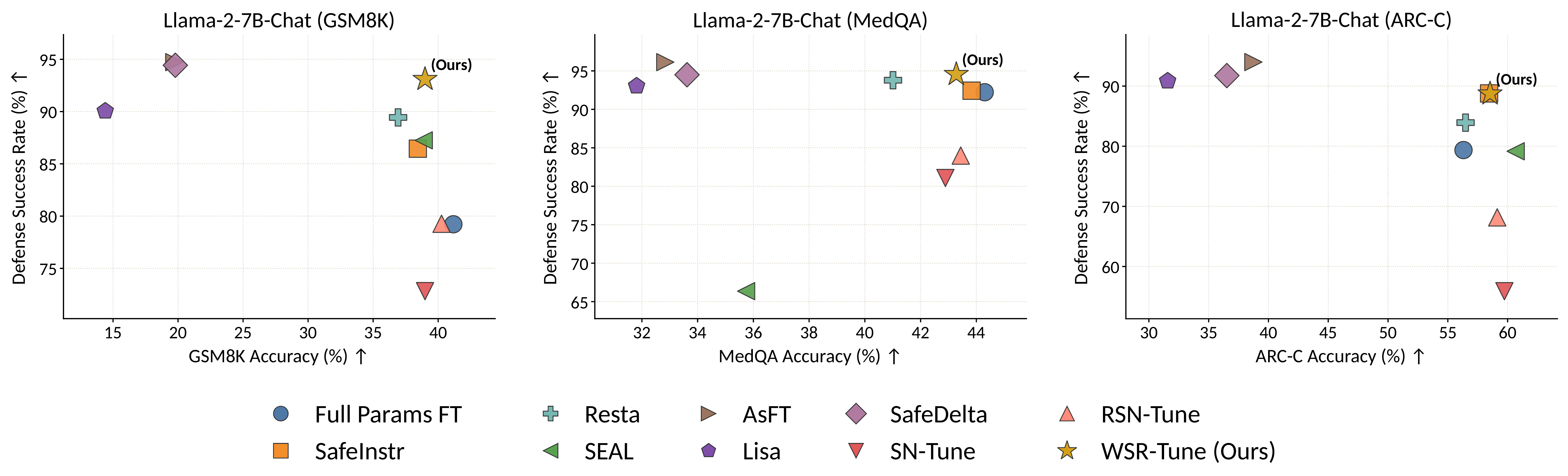}
        \vspace{-5mm}
    \caption{Comparison of safety-preserving downstream among fine-tuning methods on Llama-2-7B-Chat with GSM8K, ARC-C and MedQA.}
    \label{fig:main_all_figure_2}
   % \vspace{-2mm}
\end{figure}

\textbf{Varying downstream datasets.} 
To assess whether the benefits of WSR-Tune extend beyond mathematical reasoning, we additionally evaluate it on MedQA and ARC-C (Figure~\ref{fig:main_all_figure_2}). These tasks cover diverse settings, including medical question answering and scientific knowledge reasoning. Across all settings, WSR-Tune preserves stronger safety than standard fine-tuning while remaining competitive in downstream performance, indicating that its advantages generalize beyond a single task family.
%\textcolor{orange}{WSR-LoRA likewise achieves favorable safety--utility trade-offs among PEFT methods on both tasks. Detailed results for all full-parameter and PEFT baselines are provided in Appendix~\ref{app:detailed_main_results}. These results indicate that the advantages of safety-conditioned reparameterization generalize beyond a single task and adaptation setting.}

\begin{table*}[t]
\centering
\small
\caption{Ablation study on Llama-2-7B-Chat. Lower JB AVG indicates stronger safety, and higher Accuracy indicates better downstream performance.}
\label{tab:ablation_llama2_7b_chat}
\resizebox{0.9\textwidth}{!}{
\begin{tabular}{llccccc}
\toprule
\multicolumn{2}{c}{\textbf{Llama-2-7B-Chat}} 
& \multicolumn{2}{c}{\textbf{Safety $\downarrow$}}
& \multicolumn{2}{c}{\textbf{Downstream $\uparrow$}}
& \textbf{Overall $\uparrow$} \\
\cmidrule(lr){1-2}\cmidrule(lr){3-4}\cmidrule(lr){5-6}\cmidrule(lr){7-7}
\textbf{Category} & \textbf{Method} 
& \textbf{JB AVG} & $\boldsymbol{\Delta_{\mathrm{S}}}$ 
& \textbf{Accuracy} & $\boldsymbol{\Delta_{\mathrm{D}}}$ 
& \textbf{Overall} \\
\midrule
\multirow{3}{*}{Ablation}
& Full parameter FT
& 20.78 & 0.00 & 41.17 & 0.00 & 0.00 \\

& Original space + important mask FT
& 9.17 & -11.61 & 40.41 & -0.76 & +10.85 \\

& WSR-Tune
& 6.90 & -13.88 & 38.99 & -2.18 & \cellcolor{green!20} \textbf{+11.70} \\
\bottomrule
\end{tabular}
}
%\vspace{-4mm}
\end{table*}

\textbf{Results on base models.} We also evaluate whether the same trend holds when downstream adaptation starts from base models rather than chat or instruction-tuned models. The detailed results are provided in the Appendix~\ref{app:base_model_results}, where WSR-Tune consistently achieves the best performance. This suggests that the advantage of weight-space reparameterization is not limited to a specific alignment regime, but also extends to base-to-safe-to-downstream adaptation pipelines.

% \begin{table*}[t]
% \centering
% \small
% \caption{Ablation study on Llama-2-7B-Chat. Lower JB AVG indicates stronger safety, and higher Accuracy indicates better downstream performance.}
% \label{tab:ablation_llama2_7b_chat}
% \resizebox{0.9\textwidth}{!}{
% \begin{tabular}{llccccc}
% \toprule
% \multicolumn{2}{c}{\textbf{Llama-2-7B-Chat}} 
% & \multicolumn{2}{c}{\textbf{Safety $\downarrow$}}
% & \multicolumn{2}{c}{\textbf{Downstream $\uparrow$}}
% & \textbf{Overall $\uparrow$} \\
% \cmidrule(lr){1-2}\cmidrule(lr){3-4}\cmidrule(lr){5-6}\cmidrule(lr){7-7}
% \textbf{Category} & \textbf{Method} 
% & \textbf{JB AVG} & $\boldsymbol{\Delta_{\mathrm{S}}}$ 
% & \textbf{Accuracy} & $\boldsymbol{\Delta_{\mathrm{D}}}$ 
% & \textbf{Overall} \\
% \midrule
% \multirow{3}{*}{Ablation}
% & Full parameter FT
% & 20.78 & 0.00 & 41.17 & 0.00 & 0.00 \\

% & Original space + important mask FT
% & 9.17 & -11.61 & 40.41 & -0.76 & +10.85 \\

% & WSR-Tune
% & 6.90 & -13.88 & 38.99 & -2.18 & \cellcolor{green!20} \textbf{+11.70} \\
% \bottomrule
% \end{tabular}
% }
% \vspace{-4mm}
% \end{table*}

\textbf{Ablation study.} 
%\djh{Table 번호 순서대로 등장해야함 (현재는 Table 6이 5보다 먼저 등장중)} 
Table~\ref{tab:ablation_llama2_7b_chat} provides an ablation study on Llama-2-7B-Chat, designed to isolate the effect of each component in our method. Full-parameter fine-tuning severely degrades safety, while simply freezing important parameters in the original space improves safety only partially. Applying an importance mask in the original parameter space improves safety by freezing parameters with high safety importance, but the average ASR is still  close to 10\%. 
 The full WSR-Tune method, which applies safety-aware masking in the reparameterized space, yields the best overall balance, confirming that both components are important for the final performance gain.

\textbf{Impact of freeze ratio.} 
Table~\ref{tab:keep_ratio_ablation} analyzes the effect of the freeze ratio using Llama-2-7B-Chat. Increasing the freeze ratio to $20\%$ freezes more safety-related coefficients,  improving safety preservation while reducing the capacity for downstream adaptation. Overall, WSR-Tune remains robust across a moderate range of freeze ratios, maintaining low average ASR and competitive downstream performance. Results for freeze ratios up to $50\%$ are provided in Appendix~\ref{app:freeze_ratio}.

\begin{wraptable}{r}{0.55\columnwidth}
\vspace{-3mm}
\centering
\scriptsize
\caption{Effect of the freeze ratio in WSR-Tune.}
\label{tab:keep_ratio_ablation}
% \vspace{-3mm}
\renewcommand{\arraystretch}{0.85}
\setlength{\tabcolsep}{3pt}

\resizebox{0.55\columnwidth}{!}{
\begin{tabular}{ccccccc}
\toprule
\textbf{} & \multicolumn{5}{c}{\textbf{Safety $\downarrow$}} & \textbf{Downstream $\uparrow$} \\
\cmidrule(lr){2-6}\cmidrule(lr){7-7}
\textbf{Freeze ratio} & \textbf{Direct} & \textbf{AutoDAN} & \textbf{PAIR} & \textbf{PAP} & \textbf{AVG} & \textbf{Accuracy} \\
\midrule
1\%  & 0.00 & 0.00 & 10.38 & 16.88 & 6.82 & 38.82 \\
5\%  & 0.00 & 0.00 & 9.04  & 16.42 & 6.37 & 39.04 \\
10\% & 0.00 & 0.00 & 9.62  & 18.00 & 6.90 & 38.99 \\
15\% & 0.00 & 0.00 & 9.04  & 18.00 & 6.76 & 39.88 \\
20\% & 0.00 & 0.00 & 8.08  & 14.27 & 5.59 & 38.36 \\
%30\% & 0.00 & 0.00 & 11.15  & 21.12 & 8.07 & 40.41 \\
%40\% & 0.00 & 0.00 & 5.58  & 15.62 & 5.30 & 38.36 \\
%50\% & 0.00 & 0.00 & 6.15  & 14.08 & 5.06 & 37.76 \\
\bottomrule
\end{tabular}
}
% \vspace{-2mm}
\end{wraptable}

\textbf{Other results.} 
Appendix~\ref{app:runtime} reports runtime and memory measurements, showing that SVD-based basis construction incurs a one-time cost relative to downstream training. 
Appendix~\ref{app:v_basis} examines sensitivity to the output-side basis $V_l$. Appendix~\ref{app:additional_results} further reports results on Llama-3 models (Appendix~\ref{app:llama3_results}), adaptation from the Llama-3.1-8B base model (Appendix~\ref{app:base_model_results}), robustness when BeaverTails is used as the safety dataset and when different data fractions are used for basis construction and importance masking (Appendix~\ref{app:safety_data}), freeze-ratio results up to $50\%$ (Appendix~\ref{app:freeze_ratio}), evaluation %\jsp{evaluate -> evaluation 또는 result}\wjb{evaluation이 좋은 것 같습니다} 
on AGNEWS (Appendix~\ref{app:agnews_results}), and complete attack-wise results for the main experiments (Appendix~\ref{app:detailed_main_results}). Appendix~\ref{sec:wildguard_judge} shows that WSR-Tune's favorable safety--utility trade-off persists under an LLM-based judge (WildGuard).
% Finally, Appendix~\ref{app:baseline_compatibility} extends the experiments combining WSR-Tune with existing fine-tuning methods to Llama-2-13B-Chat.

% \textbf{Other experiments.}
% Additional experiments including further analysis across model/task combinations are deferred to the Appendix~\ref{app:detailed_main_results}.

\vspace{-1mm}

\section{Conclusion} 
\vspace{-1mm}

% In this work, we revisited safety-preserving fine-tuning in LLMs and showed that existing approaches are fundamentally constrained by operating in the original parameter space, where safety information is distributed across directions. We proposed WSR-Tune, a weight-space reparameterization framework that constructs a safety-conditioned basis, enabling downstream adaptation in a complementary subspace. By concentrating safety-relevant information into a small set of coefficients, WSR-Tune allows effective preservation of safety through simple freezing while maintaining strong downstream performance. 
% Our results suggest that safety preservation depends not only on which parameters are updated, but also on how the parameter space is structured, highlighting weight-space reparameterization as a promising direction for reliable LLM adaptation. 
In this work, we revisited safety-preserving fine-tuning in LLMs and showed that existing approaches are fundamentally constrained by operating in the original parameter space, where safety information is distributed across directions. We proposed WSR-Tune, a weight-space reparameterization framework that constructs a safety-conditioned basis, enabling downstream adaptation in a complementary subspace. By concentrating safety-relevant information into a small set of coefficients, WSR-Tune allows effective preservation of safety through simple freezing while maintaining strong downstream performance. 
While prior work highlights the entanglement of safety- and task-related information in the original parameter space, our findings suggest that changing parameter coordinates can enable more targeted protection of safety-relevant components.
Together, these results suggest that safety preservation depends not only on which parameters are updated, but also on the coordinates in which they are protected, highlighting weight-space reparameterization as a promising direction for balancing safety preservation and downstream task adaptation.

\subsubsection*{Acknowledgments}
This work was supported in part by the ``Advanced GPU Utilization Support Program'' funded by the Government of the Republic of Korea (Ministry of Science and ICT), the National Research Foundation of Korea (NRF) grants funded by the Korea government (MSIT) (RS-2026-25470983, RS-2026-25612350), and the Institute of Information \& Communications Technology Planning \& Evaluation (IITP) grant funded by the Korea government (MSIT) (RS-2024-00457882, AI Research Hub Project). Dong-Jun Han is the corresponding author.


\bibliographystyle{unsrt}
\bibliography{ref}

%%%%%%%%%%%%%%%%%%%%%%%%%%%%%%%%%%%%%%%%%%%%%%%%%%%%%%%%%%%%
\clearpage
\appendix

\section{Experimental Setup}
\label{app:experimental_setup}

All experiments are conducted on a server equipped with a single NVIDIA B200 GPU with 180GB of memory, AMD EPYC 9365 36-core processor $\times$ 2 (72 cores total), and 2.2TB of system memory. We use bf16 mixed-precision training with a maximum sequence length of 1024 tokens. For training, we use gradient accumulation to obtain an effective batch size of 16.

\subsection{Training Hyperparameters}
\label{app:hyperparameters}

\paragraph{Shared settings.}
Downstream training uses 3 epochs, an effective batch size of 16, a maximum sequence length of 1024, bf16 precision, and random seed 42. We use AdamW with cosine learning-rate decay and a gradient-clipping norm of 1.0. 

\paragraph{Full-parameter adaptation.}
Safety tuning uses a learning rate of $5\times10^{-5}$ for the Llama and Qwen models and $3\times10^{-5}$ for Gemma-2-9B-IT. Full-parameter downstream training uses a learning rate of $5\times10^{-5}$, weight decay of 0.01, and a warmup ratio of 0.1. For Gemma-2-9B-IT, the downstream learning rate is $1\times10^{-5}$. SN-Tune, RSN-Tune, and WSR-Tune target the attention projections \texttt{q}, \texttt{k}, and \texttt{v} and the MLP projections \texttt{up} and \texttt{down} in all transformer layers.

\paragraph{LoRA-based adaptation.}
Vanilla LoRA FT, AsFT, Lisa, SafeLoRA, SaLoRA, and WSR-LoRA use rank $r=16$, scaling parameter $\alpha=16$ (i.e., $\alpha/r=1$), and dropout 0.05. Adapters are applied to the same five projections (\texttt{q}, \texttt{k}, \texttt{v}, \texttt{up}, and \texttt{down}) in all transformer layers. For GSM8K, MATH, MedQA, and ARC-C, all six methods use a learning rate of $3\times10^{-4}$, weight decay of 0, and a warmup ratio of 0.03. The same settings are used for the BeaverTails-based comparison.

% Values below follow the runs reported in the current tables in RESULTS.md,
% rather than the older defaults in scripts/revision/common.sh:
% LoRA alpha=16, Lisa rho=1.0, and WSR-LoRA freeze ratio=0.3.
\paragraph{Method-specific settings.}
Table~\ref{tab:method_hyperparameters} lists the additional settings for each method. WSR-Tune uses a default freeze ratio of 10\%, whereas the reported WSR-LoRA comparisons use 30\%. WSR-LoRA uses function-preserving SVD initialization and importance-based masks on both low-rank factors, as detailed in Appendix~\ref{app:wsr_lora}.

\begin{table}[htbp]
\centering
\small
\caption{Method-specific settings used in the reported comparisons, unless explicitly varied in an ablation. Common optimization and adapter settings are described above.}
\label{tab:method_hyperparameters}
\renewcommand{\arraystretch}{1.15}
\begin{tabular}{@{}>{\raggedright\arraybackslash}p{0.23\textwidth}>{\raggedright\arraybackslash}p{0.73\textwidth}@{}}
\toprule
\textbf{Method} & \textbf{Additional settings} \\
\midrule
Full Params FT & No additional safety constraint. \\
SafeInstr & Safety examples equal to 10\% of the downstream training-set size. \\
Resta & Safety-vector scaling coefficient $\gamma=0.3$. \\
SafeDelta & Safety degradation constraint $s=0.1$. \\
SEAL & Select the top 80\% of downstream samples. \\
SN-Tune / RSN-Tune & Safety-neuron or critical-safety-neuron budget of at most 1\% of model parameters. \\
WSR-Tune & Freeze ratio $\rho=0.1$ in the reparameterized weight space. \\
\midrule
Vanilla LoRA FT & No additional safety constraint. \\
AsFT & Alignment regularization coefficient $\lambda=1.0$. \\
Lisa & Proximal coefficient $\rho_{\mathrm{Lisa}}=1.0$; alternate 100 safety-alignment steps and 900 downstream steps. \\
SafeLoRA & Cosine-similarity threshold $\tau=0.3$ for selecting updates to project. \\
SaLoRA & Safety and task subspace ranks $r_s=r_t=32$; task-specific initialization using 256 safety examples and 256 downstream examples. \\
WSR-LoRA & Freeze ratio $\rho=0.3$; importance masks on both $B$ and $\widetilde A=AU$. \\
\bottomrule
\end{tabular}
\end{table}

\section{Runtime and Computational Overhead}
\label{app:runtime}
%\kms{Rebuttal에서 측정한 Time/VRAM Cost로 업데이트}
We analyze the computational overhead of WSR-Tune. All measurements are conducted on the Llama-2-7B-Chat model with GSM8K as the downstream dataset, using a single NVIDIA B200 GPU.

\paragraph{Basis construction cost.}
In WSR-Tune, an additional preprocessing step is required to construct the safety-conditioned basis. This involves performing SVD on all target layers (32 layers, including attention projections \texttt{q}, \texttt{k}, \texttt{v} and MLP projections \texttt{up}, \texttt{down}), sharing a single SVD across projections with the same input activations. This step takes approximately 9 minutes 31 seconds and requires around 30\,GB of GPU memory. Notably, this cost is incurred only once before downstream fine-tuning.

\begin{table}[htbp]
\centering
\caption{Measured costs of WSR-Tune on Llama-2-7B-Chat with GSM8K using a single NVIDIA B200 GPU. Relative costs compare one-time basis construction with the three-epoch WSR-Tune downstream training stage.}
\label{tab:runtime_cost}
\begin{tabular}{lcc}
\toprule
Process & Time & VRAM Cost \\
\midrule
Basis construction (one time) & 9 min 31 s & 30\,GB \\
Downstream training (3 epochs) & 39 min 8 s & 96\,GB \\
Basis construction / training & $\approx 0.24$ & $\approx 0.32$ \\
\bottomrule
\end{tabular}
\end{table}

\paragraph{Training time comparison.}
For downstream adaptation, WSR-Tune requires 39 minutes 8 seconds for 3 epochs (approximately 13 minutes per epoch), with a peak GPU memory usage of 96\,GB. As shown in Table~\ref{tab:runtime_cost}, basis construction adds approximately 24\% relative to this three-epoch fine-tuning time, while requiring approximately 0.32$\times$ the GPU memory of the training stage. Also, WSR-LoRA reduces this cost: including the one-time basis construction, it requires approximately 0.32× the end-to-end wall-clock time and 0.55× the peak VRAM of WSR-Tune, and reusing the fixed basis lowers the run-specific time further to 0.18×. 

\paragraph{Summary.}
Overall, WSR-Tune introduces additional computational overhead due to basis construction and reparameterized training. However, since the basis construction is a one-time cost, its relative cost decreases over longer training runs. The method remains practical while providing an improved trade-off between safety and downstream performance.

\section{Effect of the Output-Side Basis Choice}
\label{app:v_basis}
% \kms{$V_{perm}$과 $V_{dense}$ 실험 방식을 자세히 적어야할지 아니면 아래와 같이 재현 가능한 정도로만 서술해도 괜찮을지..}

% \wjb{V = I 로 정의한 이유만 표현되어도 좋을 것 같음. 현재 수준이 적합해보인다는 생각임.}

In the main experiments, we set the output-side basis to $V_l=I$ in Equation (\ref{eq:reparameterize_weight_matrix}).
To examine the sensitivity of WSR-Tune to this choice, we additionally consider two orthogonal alternatives while keeping all other settings unchanged: a signed permutation matrix $V_{\mathrm{perm}}$, which permutes and sign-flips the output coordinates without mixing them, and a dense Haar-random orthogonal matrix $V_{\mathrm{dense}}$, which fully mixes the output coordinates.

\begin{table}[]
\centering
\caption{Effect of the output-side basis $V$ on Llama-2-7B-Chat with GSM8K.}
\label{tab:v_basis}
\begin{tabular}{lccccc}
\toprule
Method & JB AVG $\downarrow$ & $\Delta_S$ $\downarrow$ & GSM8K $\uparrow$ & $\Delta_D$ $\uparrow$ & Overall $\uparrow$ \\
\midrule
Full Params FT
& 20.78 & 0.00 & 41.17 & 0.00 & 0.00 \\
WSR-Tune ($V=V_{\mathrm{perm}}$)
& 8.38 & -12.40 & 39.95 & -1.22 & +11.18 \\
WSR-Tune ($V=V_{\mathrm{dense}}$)
& 8.21 & -12.57 & 39.65 & -1.52 & +11.05 \\
WSR-Tune ($V=I$)
& \textbf{6.90} & \textbf{-13.88} & 38.99 & -2.18 & \textbf{+11.70} \\
\bottomrule
\end{tabular}
\end{table}

As shown in Table~\ref{tab:v_basis}, all three choices of $V_l$ yield similar safety--utility trade-offs and substantially improve over full-parameter fine-tuning. Although changing $V_l$ can alter the coordinate-wise importance scores and the resulting mask, the overall performance varies only modestly.

This is consistent with the role of $V_l$ and $U_l$ in
$\widetilde W_l = V_l^\top W_l U_l$: $U_l$ is constructed from the safety activation covariance and therefore encodes safety-conditioned input directions, whereas $V_l$ only changes the orthonormal coordinates of the output dimension without introducing additional safety information. We therefore use $V_l=I$ throughout the main experiments as the simplest canonical choice that avoids introducing an arbitrary output-side rotation.

\section{Mechanistic Effect of Reparameterized Protection}
\label{app:reparam_analysis}
%\kms{Reviewer 3이 말한 "WSR의 Novelty는 어디에 있는건가?"에 대한 답변.}
We examine how safety-conditioned reparameterization changes the effect of the same importance-based protection rule, complementing the ablation in Table~\ref{tab:ablation_llama2_7b_chat}.

\paragraph{Induced update constraints.} With $V_l=I$, we have $\widetilde W_l=W_lU_l$. Let $S_{l,i}=\{j:M_l(i,j)=1\}$ denote the protected basis indices for row $i$. Freezing $\widetilde W_l(i,j)$ imposes $e_i^\top\Delta W_lu_j=0$, where $e_i$ selects the $i$-th output row and $u_j$ is the $j$-th column of $U_l$. Equivalently, 
\begin{equation}
    P_{l,i}=\sum_{j\in S_{l,i}}u_ju_j^\top,
    \qquad
    e_i^\top\Delta W_lP_{l,i}=0.
\end{equation}
Different rows can therefore preserve different subsets of the shared basis. Original-space masking is the special case $U_l=I$, which freezes individual weight entries rather than their linear combinations.

\paragraph{Two consequences.}
First, the protection budget is distributed differently. At $\rho=10\%$, 99.9\% of original weight coordinates participate in the induced constraints, and 98.2\% of all coordinates are only partially constrained: they can still change while satisfying the coupled constraints. The reparameterized mask also mixes protected and trainable entries in 99.22\% of rows and 95.74\% of columns.

Second, since $U_l$ diagonalizes $C_l=H_lH_l^\top$, the protected and complementary directions satisfy
\begin{equation}
    P_{l,i}C_l(I-P_{l,i})=0.
\end{equation}
Thus, these directions decouple under the empirical safety activation statistics used to construct the basis, a property that canonical-coordinate masking does not generally satisfy.

\paragraph{Held-out safety selectivity.}
We evaluate the masks on 256 BeaverTails safety examples and 256 GSM8K task examples, none used to construct the basis or masks. For each dataset $D$, we measure the protected activation-energy fraction $P_D$ in each mask's own basis across 160 projections (\texttt{q}, \texttt{k}, \texttt{v}, \texttt{up}, and \texttt{down} over 32 layers), excluding padding tokens. We report $R=P_{\mathrm{safe}}/P_{\mathrm{task}}$, with overall energy fractions and projection-wise ratios shown separately in Table~\ref{tab:protected_energy_overall} and ~\ref{tab:protected_energy_projection}.

\begin{table}[]
\centering
\caption{
Overall protected activation-energy fractions under a 10\% masking budget. Safety and Task denote $P_{\mathrm{safe}}$ and
$P_{\mathrm{task}}$, respectively; $R=P_{\mathrm{safe}}/P_{\mathrm{task}}$.
}
\label{tab:protected_energy_overall}
\begin{tabular}{lccc}
\toprule
\textbf{Mask space} & \textbf{Safety} & \textbf{Task} & $\boldsymbol{R}$ \\
\midrule
Original & 0.321 & 0.287 & 1.12 \\
WSR & \textbf{0.530} & 0.359 & \textbf{1.48} \\
\bottomrule
\end{tabular}
\end{table}

\begin{table}[]
\centering
\caption{
Projection-wise safety-to-task activation-energy ratios under a 10\% masking budget. Each entry reports the mean per-layer $R$ over 32 transformer layers.
}
\label{tab:protected_energy_projection}
\begin{tabular}{lccccc}
\toprule
\textbf{Mask space}
& \textbf{q} & \textbf{k} & \textbf{v}
& \textbf{up} & \textbf{down} \\
\midrule
Original & 1.08 & 1.08 & 1.08 & 1.08 & 1.26 \\
WSR & \textbf{1.36} & \textbf{1.36} & \textbf{1.41}
& \textbf{1.83} & \textbf{1.54} \\
\bottomrule
\end{tabular}
\end{table}

WSR increases the overall ratio from 1.12 to 1.48 and achieves higher ratios across all projection types. Here, $R=1$ indicates equal relative coverage of safety and task activation energy. Together with Table~\ref{tab:ablation_llama2_7b_chat}, these results support that reparameterization changes what the importance criterion selects and how protection constrains updates, without implying complete separation of safety and task representations.

\section{Detailed Formulation of WSR-LoRA}
\label{app:wsr_lora}

We provide the detailed formulation of WSR-LoRA. For clarity, we omit the layer index and the standard LoRA~\citep{lora} scaling factor; the same procedure is applied independently to each adapted linear projection. 

\paragraph{Function-preserving low-rank initialization.} Standard LoRA starts from a zero low-rank update, such that the pre-existing safety-aligned structure remains entirely in the frozen weight $W_0$. To expose a nonzero component of the aligned weight to the WSR protection mechanism, we adopt the function preserving low-rank initialization of PiSSA~\citep{pissa}. %\wjb{PiSSA work를 굳이 언급하지 않는 것은 어떨지요?} \kms{논문의 main idea가 아니고 appendix에서 다루기 때문에 언급해도 괜찮아 보입니다.}
%To expose a nonzero component of the aligned weight to the WSR protection mechanism, we use a function-preserving low-rank parameterization.

Let
\begin{equation}
    W_0=P\Lambda Q^\top
\end{equation}
be the singular value decomposition of the safety-aligned weight.
Using the leading $r$ singular components, we initialize
\begin{equation}
    B_0=P_r\Lambda_r^{1/2},
    \qquad
    A_0=\Lambda_r^{1/2}Q_r^\top,
\end{equation}
and define the frozen residual weight as
\begin{equation}
    W_{\mathrm{res}}=W_0-B_0A_0.
\end{equation}
Therefore,
\begin{equation}
    W_0=W_{\mathrm{res}}+B_0A_0.
\end{equation}

This initialization is used only to obtain a nonzero trainable factorization that exactly reproduces the original model at initialization; it does not assume that the selected singular components are themselves safety-specific. The safety-aware component of WSR-LoRA is introduced by the subsequent reparameterization and importance-based protection.

\paragraph{Safety-conditioned reparameterization.} 
%\wjb{Notation 검토가 필요합니다. 제 생각에 $\tilde A$와 관련해서 $\tilde A_0$가 잘못된 표현 같다는 생각이고, $B_0, A_0$의 표현 자체를 사용하지 않는게 더 맞는 표현 같습니다.} \kms{처음 정의부터 $A_t$로 정의하여 t는 학습단계, t=0은 initialize 시점임을 명시했습니다. }
Let $U\in\mathbb{R}^{n\times n}$ denote the safety-conditioned input basis constructed as in Section~\ref{sec:wsr-algorithm}. At downstream optimization step $t$, we reparameterize the input-side factor as
\begin{equation}
    \widetilde A_t=A_tU,
    \qquad
    A_t=\widetilde A_t U^\top.
\end{equation}
The adapted weight can then be written as
\begin{equation}
    W_t
    =
    W_{\mathrm{res}}
    +
    B_t\widetilde A_tU^\top.
\end{equation}
At initialization ($t=0$), $\widetilde A_0=A_0U$, and since $UU^\top=I$,
\begin{equation}
    W_{\mathrm{res}}
    +
    B_0\widetilde A_0U^\top
    =
    W_{\mathrm{res}}+B_0A_0
    =
    W_0.
\end{equation}
Thus, WSR-LoRA preserves the original model function at initialization while expressing the trainable input-side factor in the same safety-conditioned coordinates used by WSR-Tune.

\paragraph{Importance estimation and protection.}
During downstream adaptation, $W_{\mathrm{res}}$ and $U$ remain fixed, while $B$ and $\widetilde A$ are trainable. Following the same importance criterion as WSR-Tune, we compute
\begin{equation}
    G_B
    =
    \sum_{x\in D_{\mathrm{safe}}}
    \left|
    \frac{\partial \mathcal L(x)}{\partial B}
    \right|,
    \qquad
    G_{\widetilde A}
    =
    \sum_{x\in D_{\mathrm{safe}}}
    \left|
    \frac{\partial \mathcal L(x)}{\partial \widetilde A}
    \right|.
\end{equation}
We construct binary masks $M_B$ and $M_{\widetilde A}$ by selecting the top-$\rho$ entries according to these scores. During downstream fine-tuning, the selected safety-critical entries of $B$ and $\widetilde A$ are kept fixed at their initialized values, while only the remaining entries are updated.

Unlike full-parameter WSR-Tune, the effective low-rank update depends bilinearly on $B$ and $\widetilde A$. Therefore, factor-wise protection does not impose the exact coefficient-wise constraints of WSR-Tune; rather, WSR-LoRA transfers the same safety-conditioned protection principle to a low-rank trainable parameterization.

\newpage

\section{Additional Experimental Results}
\label{app:additional_results}

% \wjb{이 섹션을 3번째로 올리면 어떨까 싶음.}

This appendix provides additional experimental results omitted from the main text due to space constraints. 
In the main paper, Table~\ref{tab:main_all_models_delta} reports average safety changes across models. 
Here, we provide the full breakdown for each model, including attack-specific safety results for Direct, AutoDAN, PAIR, and PAP, as well as downstream performance on each evaluated dataset.

\subsection{Results on Llama-3 Models}
\label{app:llama3_results}

Table~\ref{tab:llama3_models_delta} compares safety-preserving fine-tuning methods on Llama-3.2-3B-Instruct and Llama-3.1-8B-Instruct with MATH fine-tuning. WSR-Tune improves Overall by +3.11 and +5.24 points over Full Params FT, respectively, while WSR-LoRA improves Overall by +2.35 and +0.34 points over Vanilla LoRA FT. The full attack-wise breakdown is provided in Table~\ref{tab:main_llama3_models}.

\begin{table}[htbp]
\centering
\scriptsize
\caption{Results on Llama-3.2-3B-Instruct and Llama-3.1-8B-Instruct with MATH fine-tuning. Metrics follow Table~\ref{tab:main_all_models_delta}; deltas in the upper block, including AsFT and Lisa, are relative to Full Params FT, while those in the lower block are relative to Vanilla LoRA FT. AsFT and Lisa use LoRA adapters in these experiments. Lower JB AVG and higher MATH accuracy and Overall are better.}
\label{tab:llama3_models_delta}
\resizebox{\textwidth}{!}{
\begin{tabular}{lccccc|ccccc}
\toprule
& \multicolumn{5}{c|}{\textbf{Llama-3.2-3B-Instruct}} 
& \multicolumn{5}{c}{\textbf{Llama-3.1-8B-Instruct}} \\
\cmidrule(lr){2-6} \cmidrule(lr){7-11}
& \multicolumn{2}{c}{\textbf{Safety $\downarrow$}}
& \multicolumn{2}{c}{\textbf{Downstream $\uparrow$}}
& \multicolumn{1}{c|}{\textbf{Overall} $\uparrow$}
& \multicolumn{2}{c}{\textbf{Safety $\downarrow$}}
& \multicolumn{2}{c}{\textbf{Downstream $\uparrow$}}
& \multicolumn{1}{c}{\textbf{Overall} $\uparrow$} \\
\cmidrule(lr){2-3} \cmidrule(lr){4-5} \cmidrule(lr){6-6}
\cmidrule(lr){7-8} \cmidrule(lr){9-10} \cmidrule(lr){11-11}
\textbf{Method}
& \textbf{JB AVG} & $\boldsymbol{\Delta_{\mathrm{S}}}$
& \textbf{MATH} & $\boldsymbol{\Delta_{\mathrm{D}}}$ & $\boldsymbol{\Delta_{\mathrm{Overall}}}$
& \textbf{JB AVG} & $\boldsymbol{\Delta_{\mathrm{S}}}$
& \textbf{MATH} & $\boldsymbol{\Delta_{\mathrm{D}}}$ & $\boldsymbol{\Delta_{\mathrm{Overall}}}$ \\
\midrule

Full Params FT
& 8.65 & 0.00 & 21.52 & 0.00 & 0.00
& 9.28 & 0.00 & 12.12 & 0.00 & 0.00 \\

SafeInstr
& 8.23 & -0.42 & 22.26 & +0.74 &  +1.16
& 6.21 & -3.07 & 13.98 & +1.86 &  +4.93 \\

Resta
& 8.12 & -0.53 & 20.56 & -0.96 & -0.43
& 6.74 & -2.54 & 10.56 & -1.56 & +0.98 \\

SEAL
& 8.37 & -0.28 & 21.62 & +0.10 & +0.38
& 5.45 & -3.83 & 13.18 & +1.06 & +4.89 \\

AsFT
& 7.11 & -1.54 & 7.76 & -13.76 & -12.22
& 3.36 & -5.92 & 5.66 & -6.46 & -0.54 \\

Lisa
& 6.83 & -1.82 & 8.82 & -12.70 & -10.88
& 3.52 & -5.76 & 11.32 & -0.80 & +4.96 \\

SafeDelta
& 6.48 &  -2.17 & 6.48 & -15.04 & -12.87
& 4.55 &  -4.73 & 1.18 & -10.94 & -6.21 \\

SN-Tune
& 21.84 & +13.19 & 19.36 & -2.16 & -15.35
& 54.35 & +45.07 & 13.62 & +1.50 & -43.57 \\

RSN-Tune
& 24.49 & +15.84 & 19.62 & -1.90 & -17.74
& 55.34 & +46.06 & 14.10 & +1.98 & -44.08 \\

\textbf{WSR-Tune (Ours)}
& 6.40 &  -2.25  & 22.38 &  +0.86  & \cellcolor{green!20}\textbf{+3.11}
& 5.62 &  -3.66 & 13.70 & +1.58 & \cellcolor{green!20}\textbf{+5.24} \\

\midrule

Vanilla LoRA FT
& 8.23 & 0.00 & 24.80 & 0.00 & 0.00
& 5.82 & 0.00 & 25.30 & 0.00 & 0.00 \\

SafeLoRA
& 6.26 & -1.97 & 23.04 & -1.76 & +0.21
& 3.94 & -1.88 & 22.04 & -3.26 & -1.38 \\

SaLoRA
& 9.35 & +1.12 & 22.38 & -2.42 & -3.54
& 6.56 & +0.74 & 23.20 & -2.10 & -2.84 \\

\textbf{WSR-LoRA (Ours)}
& 6.76 & -1.47 & 25.68 & +0.88 & \cellcolor{green!20}\textbf{+2.35}
& 4.18 & -1.64 & 24.00 & -1.30 & \cellcolor{green!20}\textbf{+0.34} \\

\bottomrule
\end{tabular}
}
\end{table}

\subsection{Results on Llama-3.1-8B Base}
\label{app:base_model_results}

Table~\ref{tab:llama31_8b_base_results} reports full-parameter and PEFT results when downstream adaptation starts from Llama-3.1-8B Base. Deltas are computed relative to Full Params FT for full-parameter methods and Vanilla LoRA FT for PEFT methods. WSR-Tune and WSR-LoRA achieve the best overall trade-offs in their respective adaptation regimes, showing that weight-space reparameterization generalizes to base-to-safe-to-downstream adaptation.

\begin{table*}[t]
\centering
\scriptsize
\caption{Results on Llama-3.1-8B Base with GSM8K fine-tuning. Safety is measured by ASR under Direct, AutoDAN, PAIR, and PAP, with JB AVG denoting their mean. Raw scores are percentages and deltas are percentage-point differences. Deltas in the upper comparison block, including AsFT and Lisa, are relative to Full Params FT; those in the lower comparison block are relative to Vanilla LoRA FT. AsFT and Lisa use LoRA adapters. We use $\Delta_{\mathrm{Overall}}=-\Delta_{\mathrm{S}}+\Delta_{\mathrm{D}}$. Green cells highlight the Overall changes of WSR methods.}
\label{tab:llama31_8b_base_results}
\resizebox{\textwidth}{!}{
\begin{tabular}{lccccccccc}
\toprule
 & \multicolumn{6}{c}{\textbf{Safety $\downarrow$}} & \multicolumn{2}{c}{\textbf{Downstream $\uparrow$}} & \textbf{Overall $\uparrow$} \\
\cmidrule(lr){2-7} \cmidrule(lr){8-9} \cmidrule(lr){10-10}
\textbf{Method} & \textbf{Direct} & \textbf{AutoDAN} & \textbf{PAIR} & \textbf{PAP} & \textbf{JB AVG} & $\boldsymbol{\Delta_{\mathrm{S}}}$ & \textbf{GSM8K} & $\boldsymbol{\Delta_{\mathrm{D}}}$ & $\boldsymbol{\Delta_{\mathrm{Overall}}}$ \\
\midrule
Full Params FT
& 0.00 & 0.19 & 9.04 & 26.00 & 8.81 & 0.00 & 57.32 & 0.00 & 0.00 \\
SafeInstr
& 0.00 & 0.00 & 2.88 & 23.35 & 6.56 & -2.25 & 56.94 & -0.38 & +1.87 \\
Resta
& 0.00 & 0.00 & 7.69 & 25.00 & 8.17 & -0.64 & 48.45 & -8.87 & -8.23 \\
SEAL
& 0.00 & 0.19 & 8.08 & 25.00 & 8.32 & -0.49 & 55.95 & -1.37 & -0.88 \\
AsFT
& 0.00 & 0.00 & 0.96 & 15.62 & 4.14 & -4.67 & 27.14 & -30.18 & -25.51 \\
Lisa
& 0.00 & 0.00 & 0.19 & 18.73 & 4.73 & -4.08 & 30.25 & -27.07 & -22.99 \\
SafeDelta
& 0.00 & 0.00 & 0.77 & 15.19 & 3.99 & -4.82 & 41.32 & -16.00 & -11.18 \\
SN-Tune
& 9.04 & 20.96 & 48.27 & 69.12 & 36.85 & +28.04 & 63.00 & +5.68 & -22.36 \\
RSN-Tune
& 23.27 & 12.12 & 40.38 & 38.38 & 28.54 & +19.73 & 63.53 & +6.21 & -13.52 \\
\textbf{WSR-Tune (Ours)}
& 0.00 & 0.00 & 1.35 & 16.50 & 4.46 & -4.35 & 55.12 & -2.20 & \cellcolor{green!20}\textbf{+2.15} \\
\midrule
Vanilla LoRA FT
& 1.15 & 81.73 & 38.85 & 68.12 & 47.46 & 0.00 & 57.24 & 0.00 & 0.00 \\
SafeLoRA
& 0.00 & 0.38 & 7.31 & 33.58 & 10.32 & -37.14 & 48.37 & -8.87 & +28.27 \\
SaLoRA
& 1.54 & 93.27 & 28.27 & 70.69 & 48.44 & +0.98 & 56.41 & -0.83 & -1.81 \\
\textbf{WSR-LoRA (Ours)}
& 0.00 & 5.96 & 8.08 & 27.50 & 10.38 & -37.08 & 59.29 & +2.05 & \cellcolor{green!20}\textbf{+39.13} \\
\bottomrule
\end{tabular}
}
\end{table*}

\subsection{Robustness to Safety-Data Source and Quantity}
\label{app:safety_data}

We further examine whether WSR-Tune depends strongly on the source or amount of offline safety data. In addition to the default Circuit Breakers (CB) dataset, we construct an alternative BeaverTails-based (BV) safety dataset. We screen the BeaverTails source using Llama-Guard-3-8B and match the number of safety examples to Circuit Breakers for a controlled comparison. As in the main experiments, the safety examples used by WSR consist of harmful prompts with safe refusal responses.

We evaluate three aspects of safety-data sensitivity: (i) replacing Circuit Breakers with BeaverTails as the safety source, (ii) reducing the amount of data used for basis construction or importance estimation, and (iii) using different sources for basis construction and importance estimation. All experiments use Llama-2-7B-Chat with GSM8K downstream fine-tuning.

\paragraph{Alternative safety-data source.}
Table~\ref{tab:beavertails_results} reports results when the WSR pipeline is constructed using the BeaverTails-based safety source. For full-parameter methods, $\Delta_{\mathrm{S}}$ and $\Delta_{\mathrm{D}}$ are computed relative to Full Params FT; for PEFT methods, they are computed relative to Vanilla LoRA FT.

\begin{table*}[t]
\centering
\scriptsize
\caption{Results using the BeaverTails-based safety source on Llama-2-7B-Chat with GSM8K. Safety is measured by ASR under Direct, AutoDAN, PAIR, and PAP, with JB AVG denoting their mean. Raw scores are percentages and deltas are percentage-point differences. Deltas in the upper comparison block, including AsFT and Lisa, are relative to Full Params FT; those in the lower comparison block are relative to Vanilla LoRA FT. AsFT and Lisa use LoRA adapters. We use $\Delta_{\mathrm{Overall}}=-\Delta_{\mathrm{S}}+\Delta_{\mathrm{D}}$. Green cells highlight the Overall changes of WSR methods.}
\label{tab:beavertails_results}
\resizebox{\textwidth}{!}{
\begin{tabular}{lccccccccc}
\toprule
 & \multicolumn{6}{c}{\textbf{Safety $\downarrow$}} & \multicolumn{2}{c}{\textbf{Downstream $\uparrow$}} & \textbf{Overall $\uparrow$} \\
\cmidrule(lr){2-7} \cmidrule(lr){8-9} \cmidrule(lr){10-10}
\textbf{Method} & \textbf{Direct} & \textbf{AutoDAN} & \textbf{PAIR} & \textbf{PAP} & \textbf{JB AVG} & $\boldsymbol{\Delta_{\mathrm{S}}}$ & \textbf{GSM8K} & $\boldsymbol{\Delta_{\mathrm{D}}}$ & $\boldsymbol{\Delta_{\mathrm{Overall}}}$ \\
\midrule
Full Params FT
& 0.19 & 6.73 & 34.04 & 35.69 & 19.16 & 0.00 & 41.85 & 0.00 & 0.00 \\
SafeInstr
& 0.00 & 2.88 & 19.42 & 13.38 & 8.92 & -10.24 & 40.33 & -1.52 & +8.72 \\
Resta
& 0.19 & 4.62 & 36.54 & 32.88 & 18.56 & -0.60 & 39.35 & -2.50 & -1.90 \\
SEAL
& 0.19 & 5.58 & 27.50 & 26.08 & 14.84 & -4.32 & 39.04 & -2.81 & +1.51 \\
AsFT
& 0.38 & 26.92 & 26.35 & 25.85 & 19.87 & +0.71 & 21.15 & -20.70 & -21.41 \\
Lisa
& 0.00 & 0.77 & 16.15 & 15.19 & 8.03 & -11.13 & 18.73 & -23.12 & -11.99 \\
SafeDelta
& 0.00 & 0.00 & 10.19 & 20.08 & 7.57 & -11.59 & 24.49 & -17.36 & -5.77 \\
SN-Tune
& 0.19 & 13.27 & 29.62 & 43.08 & 21.54 & +2.38 & 41.09 & -0.76 & -3.14 \\
RSN-Tune
& 0.19 & 13.65 & 35.00 & 49.73 & 24.64 & +5.48 & 41.24 & -0.61 & -6.09 \\
\textbf{WSR-Tune (Ours)}
& 0.00 & 0.19 & 13.27 & 15.27 & 7.18 & -11.98 & 41.09 & -0.76 & \cellcolor{green!20}\textbf{+11.22} \\
\midrule
Vanilla LoRA FT
& 0.19 & 0.96 & 21.92 & 32.38 & 13.86 & 0.00 & 36.62 & 0.00 & 0.00 \\
SafeLoRA
& 0.19 & 0.38 & 20.19 & 27.31 & 12.02 & -1.84 & 37.53 & +0.91 & +2.75 \\
SaLoRA
& 0.19 & 18.27 & 36.15 & 44.12 & 24.68 & +10.82 & 36.16 & -0.46 & -11.28 \\
\textbf{WSR-LoRA (Ours)}
& 0.19 & 0.58 & 18.65 & 22.65 & 10.52 & -3.34 & 36.92 & +0.30 & \cellcolor{green!20}\textbf{+3.64} \\
\bottomrule
\end{tabular}
}
\end{table*}

WSR-Tune achieves the strongest overall safety--utility trade-off among the evaluated full-parameter methods, reducing JB AVG from 19.16 to 7.18 while retaining GSM8K accuracy of 41.09. In the PEFT setting, WSR-LoRA reduces JB AVG from 13.86 to 10.52 while slightly improving downstream accuracy over Vanilla LoRA, yielding the largest Overall improvement among the evaluated PEFT methods. These results show that the benefit of WSR is
not specific to Circuit Breakers.

\paragraph{Sensitivity to basis and importance data.}
WSR uses the offline safety data for two distinct purposes: constructing the safety-conditioned activation basis and estimating coefficient importance. Table~\ref{tab:safety_data_ablation} separates these effects by varying the source and quantity of the two datasets independently. Unless otherwise specified, the full same-source safety dataset is used. ``Cross'' indicates that the basis is constructed from one safety source while importance is estimated using the other source.

\begin{table}[]
\centering
\scriptsize
\caption{
Sensitivity of WSR-Tune to the source and quantity of safety data on Llama-2-7B-Chat with GSM8K. CB and BV denote Circuit Breakers and BeaverTails, respectively. Percentages indicate the fraction of the corresponding safety source used for basis construction or importance estimation. Deltas are computed relative to Full Params FT under the corresponding safety-source setting.
}
\label{tab:safety_data_ablation}
\resizebox{\columnwidth}{!}{
\begin{tabular}{llccccc}
\toprule
\textbf{Basis data}
& \textbf{Importance data}
& \textbf{JB AVG $\downarrow$}
& \textbf{GSM8K $\uparrow$}
& $\boldsymbol{\Delta_{\mathrm{S}}}$ $\downarrow$
& $\boldsymbol{\Delta_{\mathrm{D}}}$ $\uparrow$
& \textbf{Overall $\uparrow$} \\
\midrule

\multicolumn{7}{c}{\textit{Circuit Breakers pipeline}} \\
\midrule
CB 100\% & CB 100\%
& 6.90 & 38.99 & -13.88 & -2.18 & +11.70 \\

CB 100\% & CB 50\%
& 9.09 & 40.29 & -11.69 & -0.88 & +10.81 \\

CB 100\% & CB 10\%
& 7.01 & 39.95 & -13.77 & -1.22 & +12.55 \\

CB 100\% & BV 100\%
& 8.81 & 41.02 & -11.97 & -0.15 & +11.82 \\

CB 50\% & CB 100\%
& 8.79 & 40.86 & -11.99 & -0.31 & +11.68 \\

CB 25\% & CB 100\%
& 6.08 & 40.94 & -14.70 & -0.23 & +14.47 \\

\midrule
\multicolumn{7}{c}{\textit{BeaverTails pipeline}} \\
\midrule
BV 100\% & BV 100\%
& 7.18 & 41.09 & -11.98 & -0.76 & +11.22 \\

BV 100\% & BV 50\%
& 7.88 & 40.86 & -11.28 & -0.99 & +10.29 \\

BV 100\% & BV 10\%
& 8.00 & 40.86 & -11.16 & -0.99 & +10.17 \\

BV 100\% & CB 100\%
& 7.52 & 40.86 & -11.64 & -0.99 & +10.65 \\

BV 50\% & BV 100\%
& 10.38 & 40.18 & -8.78 & -1.67 & +7.11 \\

BV 25\% & BV 100\%
& 9.23 & 38.21 & -9.93 & -3.64 & +6.29 \\

\bottomrule
\end{tabular}
}
\end{table}

The importance estimation is robust to substantial reductions in data. With the Circuit Breakers basis fixed, reducing the importance data to 50\% or 10\% yields Overall scores of +10.81 and +12.55, respectively. Similarly, with the BeaverTails basis, using only 50\% or 10\% of the importance data retains favorable trade-offs (+10.29 and +10.17). Cross-source importance estimation is also stable: using BeaverTails importance with a Circuit Breakers basis gives +11.82 Overall, while the reverse combination gives +10.65.

Basis construction is somewhat more sensitive to data reduction, particularly for BeaverTails. Nevertheless, even with only 25\% of the BeaverTails basis data, WSR-Tune reduces JB AVG from 19.16 to 9.23. For Circuit Breakers, the method remains robust even with 25\% of the basis data. Overall, these results indicate that WSR-Tune does not depend on a particular safety-data source and is especially robust to the source and quantity of the data used for importance estimation.

\subsection{Sensitivity to the Freeze Ratio}
\label{app:freeze_ratio}

Table~\ref{tab:keep_ratio_ablation_full} reports the full freeze-ratio ablation for WSR-Tune on Llama-2-7B-Chat with GSM8K fine-tuning. It extends the $1\%$--$20\%$ range shown in Table~\ref{tab:keep_ratio_ablation} to $50\%$, reporting attack-specific ASR, average ASR, and downstream accuracy.

WSR-Tune maintains low average ASR and competitive downstream accuracy across the $1\%$--$20\%$ range. Compared with the default $10\%$ setting, freezing $40\%$ or $50\%$ of coefficients yields lower average ASR with reduced GSM8K accuracy. The trend is not strictly monotonic: the $30\%$ setting yields both higher average ASR and higher downstream accuracy than the $10\%$ setting.

\begin{table}[htbp]
\centering
\small
\caption{Full freeze-ratio ablation for WSR-Tune on Llama-2-7B-Chat with GSM8K fine-tuning. Safety metrics are attack success rates (ASR), and AVG is the mean over Direct, AutoDAN, PAIR, and PAP. All values are percentages. Lower ASR and higher GSM8K accuracy are better.}
\label{tab:keep_ratio_ablation_full}
\begin{tabular}{ccccccc}
\toprule
\textbf{} & \multicolumn{5}{c}{\textbf{Safety $\downarrow$}} & \textbf{Downstream $\uparrow$} \\
\cmidrule(lr){2-6}\cmidrule(lr){7-7}
\textbf{Freeze ratio} & \textbf{Direct} & \textbf{AutoDAN} & \textbf{PAIR} & \textbf{PAP} & \textbf{AVG} & \textbf{GSM8K} \\
\midrule
1\%  & 0.00 & 0.00 & 10.38 & 16.88 & 6.82 & 38.82 \\
5\%  & 0.00 & 0.00 & 9.04  & 16.42 & 6.37 & 39.04 \\
10\% & 0.00 & 0.00 & 9.62  & 18.00 & 6.90 & 38.99 \\
15\% & 0.00 & 0.00 & 9.04  & 18.00 & 6.76 & 39.88 \\
20\% & 0.00 & 0.00 & 8.08  & 14.27 & 5.59 & 38.36 \\
30\% & 0.00 & 0.00 & 11.15  & 21.12 & 8.07 & 40.41 \\
40\% & 0.00 & 0.00 & 5.58  & 15.62 & 5.30 & 38.36 \\
50\% & 0.00 & 0.00 & 6.15  & 14.08 & 5.06 & 37.76 \\
\bottomrule
\end{tabular}
\end{table}

\subsection{Results on AGNEWS}
\label{app:agnews_results}

To evaluate downstream adaptation beyond reasoning tasks, we consider AGNEWS news classification on Llama-2-7B-Chat. Table~\ref{tab:agnews_results} reports results for selected baselines, including attack-specific ASR, average ASR, and classification accuracy. We compare AsFT and Lisa with WSR-Tune using Full Params FT as the reference, and report the four LoRA-family methods in a separate block using Vanilla LoRA FT as the reference.
% WSR-Tune reduces average ASR from 14.45\% to 5.23\% while retaining 93.78\% accuracy, compared with 94.00\% for Full Params FT, yielding $\Delta_{\mathrm{Overall}}=+9.00$. AsFT and Lisa achieve lower average ASR (4.08\% and 4.64\%, respectively), but their accuracies decrease to 91.50\% and 90.80\%. 
These results show that WSR-Tune also preserves a favorable safety--utility trade-off on a classification task. 
%In the LoRA block, WSR-LoRA increases accuracy from 92.80\% to 93.30\%, while average ASR increases from 7.52\% to 7.98\%, resulting in a small net change of $\Delta_{\mathrm{Overall}}=+0.04$.

\begin{table}[htbp]
\centering
\scriptsize
\caption{AGNEWS results on Llama-2-7B-Chat for selected baselines. Safety is measured by ASR under Direct, AutoDAN, PAIR, and PAP, with JB AVG denoting their mean; downstream performance is AGNEWS accuracy. Raw scores are percentages and deltas are percentage-point differences. Deltas in the upper block, including AsFT and Lisa, are relative to Full Params FT; those in the lower block are relative to Vanilla LoRA FT. We use $\Delta_{\mathrm{Overall}}=-\Delta_{\mathrm{S}}+\Delta_{\mathrm{D}}$, as in Table~\ref{tab:main_all_models_delta}.}
\label{tab:agnews_results}
\resizebox{\textwidth}{!}{
\begin{tabular}{lccccccccc}
\toprule
& \multicolumn{6}{c}{\textbf{Safety $\downarrow$}}
& \multicolumn{2}{c}{\textbf{Downstream $\uparrow$}}
& \textbf{Overall $\uparrow$} \\
\cmidrule(lr){2-7} \cmidrule(lr){8-9} \cmidrule(lr){10-10}
\textbf{Method}
& \textbf{Direct} & \textbf{AutoDAN} & \textbf{PAIR} & \textbf{PAP}
& \textbf{JB AVG} & $\boldsymbol{\Delta_{\mathrm{S}}}$
& \textbf{AGNEWS} & $\boldsymbol{\Delta_{\mathrm{D}}}$
& $\boldsymbol{\Delta_{\mathrm{Overall}}}$ \\
\midrule
Full Params FT
& 0.00 & 0.00 & 25.77 & 32.04 & 14.45 & 0.00 & 94.00 & 0.00 & 0.00 \\
Resta
& 0.00 & 0.00 & 14.62 & 17.54 & 8.04 & -6.41 & 93.60 & -0.40 & +6.01 \\
SEAL
& 0.00 & 0.00 & 12.69 & 14.46 & 6.79 & -7.66 & 92.50 & -1.50 & +6.16 \\
AsFT
& 0.00 & 0.00 & 5.58 & 10.73 & 4.08 & -10.37 & 91.50 & -2.50 & +7.87 \\
Lisa
& 0.00 & 0.00 & 6.92 & 11.65 & 4.64 & -9.81 & 90.80 & -3.20 & +6.61 \\
SafeDelta
& 0.00 & 0.00 & 8.65 & 23.23 & 7.97 & -6.48 & 91.10 & -2.90 & +3.58 \\
SN-Tune
& 0.96 & 55.96 & 66.15 & 89.54 & 53.15 & +38.70 & 94.00 & 0.00 & -38.70 \\
RSN-Tune
& 0.96 & 49.42 & 65.00 & 88.62 & 51.00 & +36.55 & 94.40 & +0.40 & -36.15 \\
\textbf{WSR-Tune (Ours)}
& 0.00 & 0.00 & 9.04 & 11.88 & 5.23 & -9.22 & 93.80 & -0.20 & \cellcolor{green!20}\textbf{+9.02} \\
\midrule
Vanilla LoRA FT
& 0.00 & 0.00 & 10.77 & 19.31 & 7.52 & 0.00 & 92.80 & 0.00 & 0.00 \\
SafeLoRA
& 0.00 & 0.00 & 7.50 & 15.38 & 5.72 & -1.80 & 83.20 & -9.60 & -7.80 \\
SaLoRA
& 0.00 & 0.00 & 14.23 & 24.62 & 9.71 & +2.19 & 92.70 & -0.10 & -2.29 \\
\textbf{WSR-LoRA (Ours)}
& 0.00 & 0.00 & 10.19 & 21.73 & 7.98 & +0.46 & 93.30 & +0.50 & \cellcolor{green!20}\textbf{+0.04} \\
\bottomrule
\end{tabular}
}
\end{table}

\subsection{Detailed Results for Main Experiments}
\label{app:detailed_main_results}

Table~\ref{tab:main_all_models} provides the exact attack-wise results corresponding to Table~\ref{tab:main_all_models_delta} in the main text. While Table~\ref{tab:main_all_models_delta} reports the average jailbreak score and delta-based trade-off metrics, Table~\ref{tab:main_all_models} reports the full values for Direct, AutoDAN, PAIR, and PAP, together with downstream accuracy for each model.

Table~\ref{tab:other_two_instruct_models_combined} provides the exact attack-wise results for the additional model-family experiments summarized in Table~\ref{tab:other_two_instruct_models_delta}. Similarly, Table~\ref{tab:llama2_7b_chat_medqa_arcc_delta} reports the complete results for the downstream dataset experiments visualized in Figure~\ref{fig:main_all_figure_2}, including both full-parameter and PEFT methods. These detailed tables show that the averaged improvements reported in the main text are consistent across individual attack settings and downstream datasets.

% \wjb{Table 18, 19, 20의 경우, 차라리 table 13과 같이 맞추는 편이 나아보입니다.} \wjb{Table 13, 14와 관련해서도 table 12와 같은 형태로, 수정하는 편이 어떨까 합니다. 표만 봤을 때는 WSR-Tune이 왜 좋은가 하는 부분이 잘 드러나지 않아서요.}

\begin{table*}[t]
\centering
\scriptsize
\caption{Complete results on Llama-2 models with GSM8K fine-tuning. JB AVG denotes mean ASR over the four attacks. Within each model panel, deltas are relative to Full Params FT above the divider and Vanilla LoRA FT below it. AsFT and Lisa use LoRA adapters. Raw scores are percentages and deltas are percentage-point differences, with $\Delta_{\mathrm{Overall}}=-\Delta_{\mathrm{S}}+\Delta_{\mathrm{D}}$. Green cells highlight the Overall changes of WSR methods.}
\label{tab:main_all_models}
\resizebox{\textwidth}{!}{
\begin{tabular}{lccccccccc}
\toprule
 & \multicolumn{9}{c}{\textbf{Llama-2-7B-Chat}} \\
\cmidrule(lr){2-10}
 & \multicolumn{6}{c}{\textbf{Safety $\downarrow$}} & \multicolumn{2}{c}{\textbf{Downstream $\uparrow$}} & \textbf{Overall $\uparrow$} \\
\cmidrule(lr){2-7} \cmidrule(lr){8-9} \cmidrule(lr){10-10}
\textbf{Method} & \textbf{Direct} & \textbf{AutoDAN} & \textbf{PAIR} & \textbf{PAP} & \textbf{JB AVG} & $\boldsymbol{\Delta_{\mathrm{S}}}$ & \textbf{GSM8K} & $\boldsymbol{\Delta_{\mathrm{D}}}$ & $\boldsymbol{\Delta_{\mathrm{Overall}}}$ \\
\midrule
Full Params FT
& 0.00 & 8.27 & 26.35 & 48.50 & 20.78 & 0.00 & 41.17 & 0.00 & 0.00 \\
SafeInstr
& 0.00 & 0.38 & 18.85 & 34.96 & 13.55 & -7.23 & 38.44 & -2.73 & +4.50 \\
Resta
& 0.00 & 0.38 & 15.77 & 26.08 & 10.56 & -10.22 & 36.92 & -4.25 & +5.97 \\
SEAL
& 0.00 & 4.23 & 16.92 & 29.85 & 12.75 & -8.03 & 38.89 & -2.28 & +5.75 \\
AsFT
& 0.00 & 0.00 & 8.46 & 12.62 & 5.27 & -15.51 & 19.71 & -21.46 & -5.95 \\
Lisa
& 0.00 & 0.38 & 16.35 & 23.08 & 9.95 & -10.83 & 14.40 & -26.77 & -15.94 \\
SafeDelta
& 0.00 & 0.00 & 6.92 & 15.35 & 5.57 & -15.21 & 19.79 & -21.38 & -6.17 \\
SN-Tune
& 0.00 & 17.69 & 45.58 & 45.46 & 27.18 & +6.40 & 38.99 & -2.18 & -8.58 \\
RSN-Tune
& 0.19 & 5.96 & 38.85 & 37.92 & 20.73 & -0.05 & 40.26 & -0.91 & -0.86 \\
\textbf{WSR-Tune (Ours)}
& 0.00 & 0.00 & 9.62 & 18.00 & 6.90 & -13.88 & 38.99 & -2.18 & \cellcolor{green!20}\textbf{+11.70} \\
\midrule
Vanilla LoRA FT
& 0.00 & 2.69 & 37.31 & 63.42 & 25.86 & 0.00 & 36.16 & 0.00 & 0.00 \\
SafeLoRA
& 0.00 & 0.00 & 13.46 & 19.62 & 8.27 & -17.59 & 31.08 & -5.08 & +12.51 \\
SaLoRA
& 0.00 & 5.96 & 46.15 & 70.88 & 30.75 & +4.89 & 38.74 & +2.58 & -2.31 \\
\textbf{WSR-LoRA (Ours)}
& 0.00 & 0.19 & 12.50 & 17.00 & 7.42 & -18.44 & 33.45 & -2.71 & \cellcolor{green!20}\textbf{+15.73} \\
\bottomrule
\end{tabular}
}

\vspace{1.0em}

\resizebox{\textwidth}{!}{
\begin{tabular}{lccccccccc}
\toprule
 & \multicolumn{9}{c}{\textbf{Llama-2-13B-Chat}} \\
\cmidrule(lr){2-10}
 & \multicolumn{6}{c}{\textbf{Safety $\downarrow$}} & \multicolumn{2}{c}{\textbf{Downstream $\uparrow$}} & \textbf{Overall $\uparrow$} \\
\cmidrule(lr){2-7} \cmidrule(lr){8-9} \cmidrule(lr){10-10}
\textbf{Method} & \textbf{Direct} & \textbf{AutoDAN} & \textbf{PAIR} & \textbf{PAP} & \textbf{JB AVG} & $\boldsymbol{\Delta_{\mathrm{S}}}$ & \textbf{GSM8K} & $\boldsymbol{\Delta_{\mathrm{D}}}$ & $\boldsymbol{\Delta_{\mathrm{Overall}}}$ \\
\midrule
Full Params FT
& 0.00 & 0.38 & 15.58 & 24.19 & 10.04 & 0.00 & 45.94 & 0.00 & 0.00 \\
SafeInstr
& 0.00 & 0.00 & 9.62 & 13.65 & 5.82 & -4.22 & 46.55 & +0.61 & +4.83 \\
Resta
& 0.19 & 0.00 & 15.19 & 25.96 & 10.34 & +0.30 & 44.20 & -1.74 & -2.04 \\
SEAL
& 0.00 & 0.96 & 15.19 & 24.96 & 10.28 & +0.24 & 45.19 & -0.75 & -0.99 \\
AsFT
& 0.00 & 0.00 & 1.73 & 1.73 & 0.86 & -9.18 & 28.43 & -17.51 & -8.33 \\
Lisa
& 0.00 & 0.00 & 4.62 & 0.96 & 1.39 & -8.65 & 36.24 & -9.70 & -1.05 \\
SafeDelta
& 0.00 & 0.00 & 1.54 & 2.35 & 0.97 & -9.07 & 28.43 & -17.51 & -8.44 \\
SN-Tune
& 0.00 & 24.42 & 33.08 & 58.38 & 28.97 & +18.93 & 48.22 & +2.28 & -16.65 \\
RSN-Tune
& 0.00 & 17.88 & 37.88 & 59.38 & 28.79 & +18.75 & 49.96 & +4.02 & -14.73 \\
\textbf{WSR-Tune (Ours)}
& 0.00 & 0.00 & 3.46 & 1.92 & 1.35 & -8.69 & 49.58 & +3.64 & \cellcolor{green!20}\textbf{+12.33} \\
\midrule
Vanilla LoRA FT
& 0.00 & 0.96 & 31.35 & 54.04 & 21.59 & 0.00 & 48.60 & 0.00 & 0.00 \\
SafeLoRA
& 0.00 & 0.00 & 10.58 & 3.96 & 3.64 & -17.95 & 42.30 & -6.30 & +11.65 \\
SaLoRA
& 0.00 & 0.96 & 17.88 & 52.46 & 17.82 & -3.77 & 38.44 & -10.16 & -6.39 \\
\textbf{WSR-LoRA (Ours)}
& 0.00 & 0.00 & 7.88 & 4.12 & 3.00 & -18.59 & 44.81 & -3.79 & \cellcolor{green!20}\textbf{+14.80} \\
\bottomrule
\end{tabular}
}
\end{table*}

\begin{table*}[t]
\centering
\scriptsize
\caption{Complete results on Llama-3 models with MATH fine-tuning. JB AVG denotes mean ASR over the four attacks. Within each model panel, deltas are relative to Full Params FT above the divider and Vanilla LoRA FT below it. AsFT and Lisa use LoRA adapters. Raw scores are percentages and deltas are percentage-point differences, with $\Delta_{\mathrm{Overall}}=-\Delta_{\mathrm{S}}+\Delta_{\mathrm{D}}$. Green cells highlight the Overall changes of WSR methods.}
\label{tab:main_llama3_models}
\resizebox{\textwidth}{!}{
\begin{tabular}{lccccccccc}
\toprule
 & \multicolumn{9}{c}{\textbf{Llama-3.2-3B-Instruct}} \\
\cmidrule(lr){2-10}
 & \multicolumn{6}{c}{\textbf{Safety $\downarrow$}} & \multicolumn{2}{c}{\textbf{Downstream $\uparrow$}} & \textbf{Overall $\uparrow$} \\
\cmidrule(lr){2-7} \cmidrule(lr){8-9} \cmidrule(lr){10-10}
\textbf{Method} & \textbf{Direct} & \textbf{AutoDAN} & \textbf{PAIR} & \textbf{PAP} & \textbf{JB AVG} & $\boldsymbol{\Delta_{\mathrm{S}}}$ & \textbf{MATH} & $\boldsymbol{\Delta_{\mathrm{D}}}$ & $\boldsymbol{\Delta_{\mathrm{Overall}}}$ \\
\midrule
Full Params FT
& 0.19 & 0.00 & 5.77 & 28.62 & 8.65 & 0.00 & 21.52 & 0.00 & 0.00 \\
SafeInstr
& 0.19 & 0.00 & 6.54 & 26.19 & 8.23 & -0.42 & 22.26 & +0.74 & +1.16 \\
Resta
& 0.19 & 0.00 & 7.31 & 24.96 & 8.12 & -0.53 & 20.56 & -0.96 & -0.43 \\
SEAL
& 0.00 & 0.00 & 7.31 & 26.15 & 8.37 & -0.28 & 21.62 & +0.10 & +0.38 \\
AsFT
& 0.00 & 0.00 & 5.19 & 23.23 & 7.11 & -1.54 & 7.76 & -13.76 & -12.22 \\
Lisa
& 0.00 & 0.00 & 5.96 & 21.35 & 6.83 & -1.82 & 8.82 & -12.70 & -10.88 \\
SafeDelta
& 0.00 & 0.00 & 5.19 & 20.73 & 6.48 & -2.17 & 6.48 & -15.04 & -12.87 \\
SN-Tune
& 0.38 & 0.00 & 18.46 & 68.50 & 21.84 & +13.19 & 19.36 & -2.16 & -15.35 \\
RSN-Tune
& 0.96 & 0.00 & 26.35 & 70.65 & 24.49 & +15.84 & 19.62 & -1.90 & -17.74 \\
\textbf{WSR-Tune (Ours)}
& 0.00 & 0.00 & 4.23 & 21.35 & 6.40 & -2.25 & 22.38 & +0.86 & \cellcolor{green!20}\textbf{+3.11} \\
\midrule
Vanilla LoRA FT
& 0.00 & 0.00 & 7.12 & 25.81 & 8.23 & 0.00 & 24.80 & 0.00 & 0.00 \\
SafeLoRA
& 0.00 & 0.00 & 6.15 & 18.88 & 6.26 & -1.97 & 23.04 & -1.76 & +0.21 \\
SaLoRA
& 0.19 & 0.00 & 7.50 & 29.69 & 9.35 & +1.12 & 22.38 & -2.42 & -3.54 \\
\textbf{WSR-LoRA (Ours)}
& 0.00 & 0.00 & 4.81 & 22.23 & 6.76 & -1.47 & 25.68 & +0.88 & \cellcolor{green!20}\textbf{+2.35} \\
\bottomrule
\end{tabular}
}

\vspace{1.0em}

\resizebox{\textwidth}{!}{
\begin{tabular}{lccccccccc}
\toprule
 & \multicolumn{9}{c}{\textbf{Llama-3.1-8B-Instruct}} \\
\cmidrule(lr){2-10}
 & \multicolumn{6}{c}{\textbf{Safety $\downarrow$}} & \multicolumn{2}{c}{\textbf{Downstream $\uparrow$}} & \textbf{Overall $\uparrow$} \\
\cmidrule(lr){2-7} \cmidrule(lr){8-9} \cmidrule(lr){10-10}
\textbf{Method} & \textbf{Direct} & \textbf{AutoDAN} & \textbf{PAIR} & \textbf{PAP} & \textbf{JB AVG} & $\boldsymbol{\Delta_{\mathrm{S}}}$ & \textbf{MATH} & $\boldsymbol{\Delta_{\mathrm{D}}}$ & $\boldsymbol{\Delta_{\mathrm{Overall}}}$ \\
\midrule
Full Params FT
& 0.00 & 0.00 & 4.04 & 33.08 & 9.28 & 0.00 & 12.12 & 0.00 & 0.00 \\
SafeInstr
& 0.00 & 0.00 & 1.35 & 23.50 & 6.21 & -3.07 & 13.98 & +1.86 & +4.93 \\
Resta
& 0.00 & 0.00 & 2.31 & 24.65 & 6.74 & -2.54 & 10.56 & -1.56 & +0.98 \\
SEAL
& 0.00 & 0.00 & 1.46 & 20.35 & 5.45 & -3.83 & 13.18 & +1.06 & +4.89 \\
AsFT
& 0.00 & 0.00 & 0.38 & 13.08 & 3.36 & -5.92 & 5.66 & -6.46 & -0.54 \\
Lisa
& 0.00 & 0.00 & 0.77 & 13.31 & 3.52 & -5.76 & 11.32 & -0.80 & +4.96 \\
SafeDelta
& 0.00 & 0.00 & 0.38 & 17.81 & 4.55 & -4.73 & 1.18 & -10.94 & -6.21 \\
SN-Tune
& 1.73 & 74.23 & 57.50 & 83.92 & 54.35 & +45.07 & 13.62 & +1.50 & -43.57 \\
RSN-Tune
& 6.54 & 61.35 & 68.85 & 84.62 & 55.34 & +46.06 & 14.10 & +1.98 & -44.08 \\
\textbf{WSR-Tune (Ours)}
& 0.00 & 0.00 & 1.15 & 21.31 & 5.62 & -3.66 & 13.70 & +1.58 & \cellcolor{green!20}\textbf{+5.24} \\
\midrule
Vanilla LoRA FT
& 0.00 & 0.00 & 1.54 & 21.73 & 5.82 & 0.00 & 25.30 & 0.00 & 0.00 \\
SafeLoRA
& 0.00 & 0.00 & 0.58 & 15.19 & 3.94 & -1.88 & 22.04 & -3.26 & -1.38 \\
SaLoRA
& 0.00 & 0.00 & 0.58 & 25.65 & 6.56 & +0.74 & 23.20 & -2.10 & -2.84 \\
\textbf{WSR-LoRA (Ours)}
& 0.00 & 0.00 & 0.58 & 16.15 & 4.18 & -1.64 & 24.00 & -1.30 & \cellcolor{green!20}\textbf{+0.34} \\
\bottomrule
\end{tabular}
}
\end{table*}

\begin{table*}[t]
\centering
\scriptsize
\caption{Complete results on Qwen-2.5-7B-Instruct and Gemma-2-9B-IT with GSM8K fine-tuning. Safety is measured by ASR under Direct, AutoDAN, PAIR, and PAP, with JB AVG denoting their mean. Raw scores are percentages and deltas are percentage-point differences. Deltas in the upper comparison block, including AsFT and Lisa, are relative to Full Params FT; those in the lower comparison block are relative to Vanilla LoRA FT. AsFT and Lisa use LoRA adapters. We use $\Delta_{\mathrm{Overall}}=-\Delta_{\mathrm{S}}+\Delta_{\mathrm{D}}$. Green cells highlight the Overall changes of WSR methods.}
\label{tab:other_two_instruct_models_combined}
% Values follow tem_csv.txt; deltas are computed from the reported AVG and accuracy.
\resizebox{\textwidth}{!}{
\begin{tabular}{lccccccccc}
\toprule
 & \multicolumn{9}{c}{\textbf{Qwen-2.5-7B-Instruct}} \\
\cmidrule(lr){2-10}
 & \multicolumn{6}{c}{\textbf{Safety $\downarrow$}} & \multicolumn{2}{c}{\textbf{Downstream $\uparrow$}} & \textbf{Overall $\uparrow$} \\
\cmidrule(lr){2-7} \cmidrule(lr){8-9} \cmidrule(lr){10-10}
\textbf{Method} & \textbf{Direct} & \textbf{AutoDAN} & \textbf{PAIR} & \textbf{PAP} & \textbf{JB AVG} & $\boldsymbol{\Delta_{\mathrm{S}}}$ & \textbf{GSM8K} & $\boldsymbol{\Delta_{\mathrm{D}}}$ & $\boldsymbol{\Delta_{\mathrm{Overall}}}$ \\
\midrule
Full Params FT
& 0.00 & 0.00 & 2.31 & 12.15 & 3.62 & 0.00 & 67.32 & 0.00 & 0.00 \\
SafeInstr
& 0.00 & 0.00 & 4.81 & 21.81 & 6.66 & +3.04 & 67.70 & +0.38 & -2.66 \\
Resta
& 0.00 & 0.00 & 0.96 & 6.31 & 1.82 & -1.80 & 66.87 & -0.45 & +1.35 \\
SEAL
& 0.00 & 0.00 & 3.85 & 13.50 & 4.34 & +0.72 & 70.05 & +2.73 & +2.01 \\
AsFT
& 0.00 & 0.00 & 2.12 & 12.77 & 3.72 & +0.10 & 71.57 & +4.25 & +4.15 \\
Lisa
& 0.00 & 0.00 & 2.88 & 13.77 & 4.16 & +0.54 & 70.89 & +3.57 & +3.03 \\
SafeDelta
& 0.00 & 0.00 & 1.73 & 13.54 & 3.82 & +0.20 & 66.87 & -0.45 & -0.65 \\
SN-Tune
& 1.15 & 10.96 & 53.08 & 85.00 & 37.55 & +33.93 & 69.52 & +2.20 & -31.73 \\
RSN-Tune
& 1.35 & 10.19 & 58.08 & 83.62 & 38.31 & +34.69 & 71.34 & +4.02 & -30.67 \\
\textbf{WSR-Tune (Ours)}
& 0.00 & 0.00 & 1.35 & 9.38 & 2.68 & -0.94 & 70.81 & +3.49 & \cellcolor{green!20}\textbf{+4.43} \\
\midrule
Vanilla LoRA FT
& 0.00 & 0.00 & 3.85 & 8.58 & 3.11 & 0.00 & 71.19 & 0.00 & 0.00 \\
SafeLoRA
& 0.00 & 0.00 & 2.69 & 10.26 & 3.24 & +0.13 & 73.74 & +2.55 & +2.42 \\
SaLoRA
& 0.00 & 0.00 & 5.19 & 12.42 & 4.40 & +1.29 & 70.66 & -0.53 & -1.82 \\
\textbf{WSR-LoRA (Ours)}
& 0.00 & 0.00 & 2.12 & 9.88 & 3.00 & -0.11 & 73.54 & +2.35 & \cellcolor{green!20}\textbf{+2.46} \\
\bottomrule
\end{tabular}
}

\vspace{1.0em}

\resizebox{\textwidth}{!}{
\begin{tabular}{lccccccccc}
\toprule
 & \multicolumn{9}{c}{\textbf{Gemma-2-9B-IT}} \\
\cmidrule(lr){2-10}
 & \multicolumn{6}{c}{\textbf{Safety $\downarrow$}} & \multicolumn{2}{c}{\textbf{Downstream $\uparrow$}} & \textbf{Overall $\uparrow$} \\
\cmidrule(lr){2-7} \cmidrule(lr){8-9} \cmidrule(lr){10-10}
\textbf{Method} & \textbf{Direct} & \textbf{AutoDAN} & \textbf{PAIR} & \textbf{PAP} & \textbf{JB AVG} & $\boldsymbol{\Delta_{\mathrm{S}}}$ & \textbf{GSM8K} & $\boldsymbol{\Delta_{\mathrm{D}}}$ & $\boldsymbol{\Delta_{\mathrm{Overall}}}$ \\
\midrule
Full Params FT
& 0.00 & 0.38 & 0.96 & 20.15 & 5.37 & 0.00 & 69.75 & 0.00 & 0.00 \\
SafeInstr
& 0.00 & 2.12 & 3.46 & 28.31 & 8.47 & +3.10 & 70.89 & +1.14 & -1.96 \\
Resta
& 0.00 & 0.00 & 1.35 & 16.96 & 4.58 & -0.79 & 57.85 & -11.90 & -11.11 \\
SEAL
& 0.00 & 0.38 & 0.77 & 19.61 & 5.19 & -0.18 & 68.31 & -1.44 & -1.26 \\
AsFT
& 0.00 & 0.00 & 0.77 & 18.31 & 4.77 & -0.60 & 36.62 & -33.13 & -32.53 \\
Lisa
& 0.00 & 0.00 & 0.19 & 11.92 & 3.03 & -2.34 & 65.28 & -4.47 & -2.13 \\
SafeDelta
& 0.00 & 0.00 & 0.96 & 17.12 & 4.52 & -0.85 & 64.37 & -5.38 & -4.53 \\
SN-Tune
& 0.00 & 50.38 & 41.54 & 55.54 & 36.87 & +31.50 & 74.37 & +4.62 & -26.88 \\
RSN-Tune
& 0.38 & 64.23 & 39.80 & 60.18 & 41.15 & +35.78 & 71.57 & +1.82 & -33.96 \\
\textbf{WSR-Tune (Ours)}
& 0.00 & 0.00 & 0.77 & 19.19 & 4.99 & -0.38 & 70.81 & +1.06 & \cellcolor{green!20}\textbf{+1.44} \\
\midrule
Vanilla LoRA FT
& 0.00 & 14.23 & 7.50 & 29.96 & 12.92 & 0.00 & 70.74 & 0.00 & 0.00 \\
SafeLoRA
& 0.19 & 0.00 & 1.35 & 18.23 & 4.94 & -7.98 & 69.29 & -1.45 & +6.53 \\
SaLoRA
& 0.00 & 1.92 & 10.38 & 32.96 & 11.32 & -1.60 & 68.23 & -2.51 & -0.91 \\
\textbf{WSR-LoRA (Ours)}
& 0.00 & 0.77 & 1.35 & 18.69 & 5.20 & -7.72 & 70.81 & +0.07 & \cellcolor{green!20}\textbf{+7.79} \\
\bottomrule
\end{tabular}
}
\end{table*}

\begin{table*}[t]
\centering
\scriptsize
\caption{Complete results on Llama-2-7B-Chat with MedQA and ARC-C fine-tuning. Safety is measured by ASR under Direct, AutoDAN, PAIR, and PAP, with JB AVG denoting their mean. Raw scores are percentages and deltas are percentage-point differences. Deltas in the upper comparison block, including AsFT and Lisa, are relative to Full Params FT; those in the lower comparison block are relative to Vanilla LoRA FT. AsFT and Lisa use LoRA adapters. We use $\Delta_{\mathrm{Overall}}=-\Delta_{\mathrm{S}}+\Delta_{\mathrm{D}}$. Green cells highlight the Overall changes of WSR methods.}
\label{tab:llama2_7b_chat_medqa_arcc_delta}
% Values follow tem_csv.txt; deltas are computed from the reported AVG and accuracy.
\resizebox{\textwidth}{!}{
\begin{tabular}{lccccccccc}
\toprule
 & \multicolumn{9}{c}{\textbf{MedQA}} \\
\cmidrule(lr){2-10}
 & \multicolumn{6}{c}{\textbf{Safety $\downarrow$}} & \multicolumn{2}{c}{\textbf{Downstream $\uparrow$}} & \textbf{Overall $\uparrow$} \\
\cmidrule(lr){2-7} \cmidrule(lr){8-9} \cmidrule(lr){10-10}
\textbf{Method} & \textbf{Direct} & \textbf{AutoDAN} & \textbf{PAIR} & \textbf{PAP} & \textbf{JB AVG} & $\boldsymbol{\Delta_{\mathrm{S}}}$ & \textbf{MedQA} & $\boldsymbol{\Delta_{\mathrm{D}}}$ & $\boldsymbol{\Delta_{\mathrm{Overall}}}$ \\
\midrule
Full Params FT
& 0.00 & 0.00 & 10.00 & 21.08 & 7.77 & 0.00 & 44.30 & 0.00 & 0.00 \\
SafeInstr
& 0.00 & 0.00 & 10.00 & 20.27 & 7.57 & -0.20 & 43.83 & -0.47 & -0.27 \\
Resta
& 0.00 & 0.00 & 7.50 & 17.35 & 6.21 & -1.56 & 41.01 & -3.29 & -1.73 \\
SEAL
& 0.38 & 13.27 & 45.58 & 75.31 & 33.64 & +25.87 & 35.74 & -8.56 & -34.43 \\
AsFT
& 0.00 & 0.00 & 5.19 & 10.23 & 3.86 & -3.91 & 32.84 & -11.46 & -7.55 \\
Lisa
& 0.00 & 0.00 & 12.12 & 15.73 & 6.96 & -0.81 & 31.81 & -12.49 & -11.68 \\
SafeDelta
& 0.00 & 0.00 & 6.54 & 15.46 & 5.50 & -2.27 & 33.62 & -10.68 & -8.41 \\
SN-Tune
& 0.19 & 0.96 & 25.00 & 49.58 & 18.93 & +11.16 & 42.89 & -1.41 & -12.57 \\
RSN-Tune
& 0.00 & 0.58 & 22.31 & 41.15 & 16.01 & +8.24 & 43.44 & -0.86 & -9.10 \\
\textbf{WSR-Tune (Ours)}
& 0.00 & 0.00 & 7.12 & 14.69 & 5.45 & -2.32 & 43.28 & -1.02 & \cellcolor{green!20}\textbf{+1.30} \\
\midrule
Vanilla LoRA FT
& 0.00 & 0.19 & 23.46 & 40.81 & 16.12 & 0.00 & 45.95 & 0.00 & 0.00 \\
SafeLoRA
& 0.00 & 0.19 & 20.38 & 38.77 & 14.84 & -1.28 & 45.95 & 0.00 & +1.28 \\
SaLoRA
& 0.00 & 1.15 & 29.23 & 69.54 & 24.98 & +8.86 & 42.11 & -3.84 & -12.70 \\
\textbf{WSR-LoRA (Ours)}
& 0.00 & 0.00 & 13.65 & 26.42 & 10.02 & -6.10 & 45.17 & -0.78 & \cellcolor{green!20}\textbf{+5.32} \\
\bottomrule
\end{tabular}
}

\vspace{1.0em}

\resizebox{\textwidth}{!}{
\begin{tabular}{lccccccccc}
\toprule
 & \multicolumn{9}{c}{\textbf{ARC-C}} \\
\cmidrule(lr){2-10}
 & \multicolumn{6}{c}{\textbf{Safety $\downarrow$}} & \multicolumn{2}{c}{\textbf{Downstream $\uparrow$}} & \textbf{Overall $\uparrow$} \\
\cmidrule(lr){2-7} \cmidrule(lr){8-9} \cmidrule(lr){10-10}
\textbf{Method} & \textbf{Direct} & \textbf{AutoDAN} & \textbf{PAIR} & \textbf{PAP} & \textbf{JB AVG} & $\boldsymbol{\Delta_{\mathrm{S}}}$ & \textbf{ARC-C} & $\boldsymbol{\Delta_{\mathrm{D}}}$ & $\boldsymbol{\Delta_{\mathrm{Overall}}}$ \\
\midrule
Full Params FT
& 0.00 & 0.00 & 31.54 & 51.00 & 20.64 & 0.00 & 56.31 & 0.00 & 0.00 \\
SafeInstr
& 0.00 & 0.00 & 22.88 & 22.00 & 11.22 & -9.42 & 58.45 & +2.14 & +11.56 \\
Resta
& 0.00 & 0.00 & 25.19 & 39.12 & 16.08 & -4.56 & 56.48 & +0.17 & +4.73 \\
SEAL
& 0.00 & 0.58 & 29.23 & 53.54 & 20.84 & +0.20 & 60.67 & +4.36 & +4.16 \\
AsFT
& 0.00 & 0.00 & 5.96 & 18.08 & 6.01 & -14.63 & 38.74 & -17.57 & -2.94 \\
Lisa
& 0.00 & 0.00 & 9.42 & 27.23 & 9.16 & -11.48 & 31.57 & -24.74 & -13.26 \\
SafeDelta
& 0.00 & 0.00 & 10.77 & 22.27 & 8.26 & -12.38 & 36.52 & -19.79 & -7.41 \\
SN-Tune
& 0.00 & 64.42 & 30.77 & 81.23 & 44.11 & +23.47 & 59.73 & +3.42 & -20.05 \\
RSN-Tune
& 0.19 & 18.85 & 32.88 & 75.38 & 31.83 & +11.19 & 59.13 & +2.82 & -8.37 \\
\textbf{WSR-Tune (Ours)}
& 0.00 & 0.00 & 18.27 & 26.88 & 11.29 & -9.35 & 58.53 & +2.22 & \cellcolor{green!20}\textbf{+11.57} \\
\midrule
Vanilla LoRA FT
& 0.00 & 0.00 & 20.00 & 36.23 & 14.06 & 0.00 & 61.09 & 0.00 & 0.00 \\
SafeLoRA
& 0.00 & 0.00 & 18.46 & 33.12 & 12.90 & -1.16 & 61.95 & +0.86 & +2.02 \\
SaLoRA
& 0.00 & 0.38 & 21.35 & 36.85 & 14.64 & +0.58 & 62.97 & +1.88 & +1.30 \\
\textbf{WSR-LoRA (Ours)}
& 0.00 & 0.00 & 18.85 & 29.69 & 12.13 & -1.93 & 61.57 & +0.48 & \cellcolor{green!20}\textbf{+2.41} \\
\bottomrule
\end{tabular}
}
\end{table*}

\clearpage

\section{Visualization of Baseline Compatibility Results}
\label{app:baseline_plus_wsr_visualization}

\begin{figure}[h]
    \centering
    \includegraphics[width=0.80\textwidth]{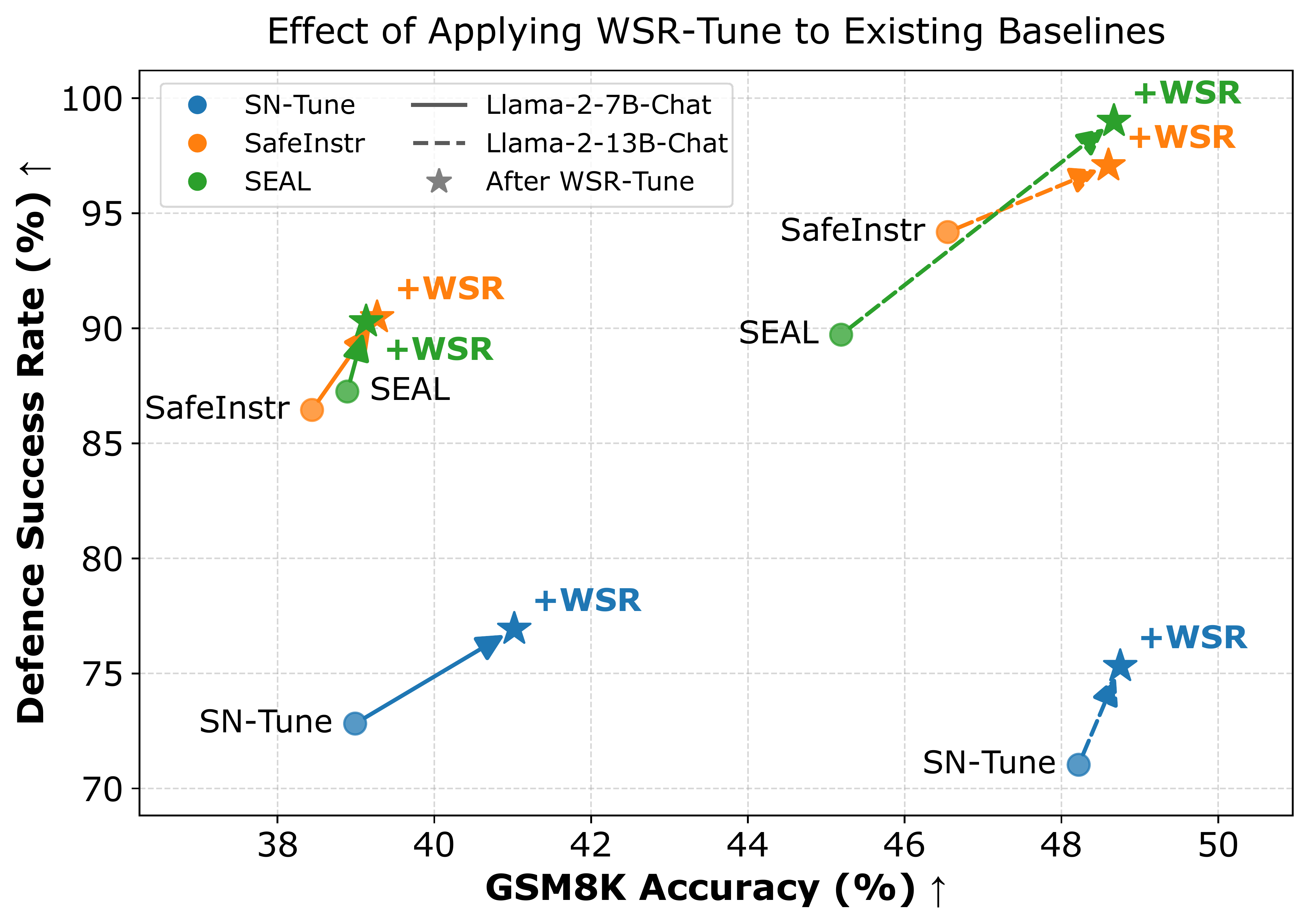}
    \caption{Visualization of the results in Table~\ref{tab:baseline_plus_warp}. Each arrow indicates the change from an existing baseline to its WSR-Tune-enhanced variant, where ``+WSR'' denotes applying WSR-Tune on top of the corresponding baseline. The x-axis denotes GSM8K accuracy, and the y-axis denotes defense success rate. Moving toward the upper-right direction indicates a better safety--downstream trade-off.}
    \label{fig:baseline_plus_wsr}
\end{figure}

Figure~\ref{fig:baseline_plus_wsr} visualizes the results in Table~\ref{tab:baseline_plus_warp}.
Each point represents the safety-downstream trade-off of an existing baseline, and each arrow indicates the change after applying WSR-Tune. The figure shows consistent improvements for SN-Tune, SafeInstr, and SEAL on Llama-2-7B-Chat and Llama-2-13B-Chat.

\FloatBarrier

\section{Robustness of the Safety Metric: An LLM-Based Refusal Judge}
\label{sec:wildguard_judge}

% \wjb{Refusal Keyword 보완을 위해 LLM-based Refusal Judge를 사용했다는 표현은 사용하지 않는 편이 낫다고 생각합니다. 그리고 LLM-based refusal judge 보다는 LLM-based classifier라는 표현이 좋지 않나 싶습니다. refusal judge라는 표현이 safety 보다는 refusal 만 estimate하는 것 같은 인상을 주기 때문입니다.} \kms{동의합니다. refusal keyword 보완이 아닌 llm classifier로도 측정을 해봤다. 라는 느낌으로 수정하겠습니다. (지금 보니 refusal keyword가 부족한 성능 평가 방식이라는 인상을 줄 수 있을 것 같습니다.}

We additionally evaluate the generated responses using WildGuard~\citep{wildguard}, an LLM-based classifier, on Llama-2-7B-Chat and Llama-2-13B-Chat after GSM8K fine-tuning. We use WildGuard's response-refusal predictions, counting a non-refusal response as a successful attack. This evaluation uses the same attack prompts, test examples, system prompt, and generated responses as the main experiments.

Table~\ref{tab:wildguard_refusal_delta} summarizes the results. WSR-Tune achieves the largest overall improvement among the compared methods. These results are consistent with the favorable safety--utility trade-off observed in our main experiments, showing that this trend also holds under LLM-based classification.
% All safety numbers reported so far use refusal keyword matching, which labels a response as a successful attack when it does not begin with a refusal phrase. A natural concern is that this lexical heuristic, rather than the methods themselves, drives the comparison. To test this, we evaluate the same generations with WildGuard~\citep{wildguard}, an open safety classifier LLM, and use its judgment of whether the response constitutes a refusal. The attacks, the test cases, the system prompt, and the generations are all unchanged; only the judge is replaced, so any difference is attributable to the metric alone.

% Table~\ref{tab:wildguard_refusal_delta} reports the result. The ordering of baselines is essentially unchanged and WSR-Tune attains the best safety--utility trade-off, with $\Delta_{\mathrm{Overall}}$ of $+8.02$ on Llama-2-7B-Chat and $+10.82$ on Llama-2-13B-Chat. These results indicate that the main refusal-based comparison is not an artifact of lexical keyword matching: replacing the heuristic with a model-based refusal classifier preserves the qualitative safety--utility pattern observed for WSR-Tune.

\begin{table*}[t]
\centering
\scriptsize
\caption{Comparison under an \textbf{LLM-based refusal judge} (WildGuard) instead of refusal-keyword matching, on Llama-2-7B-Chat and Llama-2-13B-Chat with GSM8K fine-tuning. The generations, attacks, and system prompt are identical to Table~\ref{tab:main_all_models_delta}; only the judge is replaced. 
Safety is measured by Jailbreak (JB) AVG (averaged over Direct, AutoDAN, PAIR, and PAP; lower is better); downstream performance is task accuracy (higher is better). We report changes relative to Full Params FT: $\Delta_{\mathrm{S}}$ (safety, lower is better) and $\Delta_{\mathrm{D}}$ (downstream, higher is better), and $\Delta_{\mathrm{Overall}} = -\Delta_{\mathrm{S}} + \Delta_{\mathrm{D}}$. Raw scores are percentages and deltas are percentage-point differences. Green cells highlight the Overall changes of WSR-Tune.}
\label{tab:wildguard_refusal_delta}
\resizebox{\textwidth}{!}{
\begin{tabular}{lccccc|ccccc}
\toprule
 & \multicolumn{5}{c|}{\textbf{Llama-2-7B-Chat}} & \multicolumn{5}{c}{\textbf{Llama-2-13B-Chat}} \\
\cmidrule(lr){2-6} \cmidrule(lr){7-11}
 & \multicolumn{2}{c}{\textbf{Safety $\downarrow$}} & \multicolumn{2}{c}{\textbf{Downstream $\uparrow$}} & \multicolumn{1}{c|}{\textbf{Overall $\uparrow$}} & \multicolumn{2}{c}{\textbf{Safety $\downarrow$}} & \multicolumn{2}{c}{\textbf{Downstream $\uparrow$}} & \textbf{Overall $\uparrow$} \\
\cmidrule(lr){2-3} \cmidrule(lr){4-5} \cmidrule(lr){6-6} \cmidrule(lr){7-8} \cmidrule(lr){9-10} \cmidrule(lr){11-11}
\textbf{Method} & \textbf{JB AVG} & $\boldsymbol{\Delta_{\mathrm{S}}}$ & \textbf{GSM8K} & $\boldsymbol{\Delta_{\mathrm{D}}}$ & $\boldsymbol{\Delta_{\mathrm{Overall}}}$ & \textbf{JB AVG} & $\boldsymbol{\Delta_{\mathrm{S}}}$ & \textbf{GSM8K} & $\boldsymbol{\Delta_{\mathrm{D}}}$ & $\boldsymbol{\Delta_{\mathrm{Overall}}}$ \\
\midrule
Full Params FT
& 17.64 & 0.00 & 41.17 & 0.00 & 0.00
& 8.64 & 0.00 & 45.94 & 0.00 & 0.00 \\
SafeInstr
& 15.26 & -2.38 & 38.44 & -2.73 & -0.35
& 5.82 & -2.82 & 46.55 & +0.61 & +3.43 \\
Resta
& 10.46 & -7.18 & 36.92 & -4.25 & +2.93
& 8.33 & -0.31 & 44.20 & -1.74 & -1.43 \\
SEAL
& 12.12 & -5.52 & 38.89 & -2.28 & +3.24
& 9.08 & +0.44 & 45.19 & -0.75 & -1.19 \\
AsFT
& 5.29 & -12.35 & 19.71 & -21.46 & -9.11
& 0.98 & -7.66 & 28.43 & -17.51 & -9.85 \\
Lisa
& 8.47 & -9.17 & 14.40 & -26.77 & -17.60
& 1.53 & -7.11 & 36.24 & -9.70 & -2.59 \\
SafeDelta
& 5.95 & -11.69 & 19.79 & -21.38 & -9.69
& 1.01 & -7.63 & 28.43 & -17.51 & -9.88 \\
SN-Tune
& 25.16 & +7.52 & 38.99 & -2.18 & -9.70
& 21.43 & +12.79 & 48.22 & +2.28 & -10.51 \\
RSN-Tune
& 20.64 & +3.00 & 40.26 & -0.91 & -3.91
& 21.18 & +12.54 & 49.96 & +4.02 & -8.52 \\
\textbf{WSR-Tune (Ours)}
& 7.44 & -10.20 & 38.99 & -2.18 & \cellcolor{green!20}\textbf{+8.02}
& 1.46 & -7.18 & 49.58 & +3.64 & \cellcolor{green!20}\textbf{+10.82} \\
\bottomrule
\end{tabular}
}
\end{table*}

\section{Comparison with Null space projection in Original Space}
\label{app:actsved}
% \kms{Rebuttal에서 설명한 내용 설명 및 실험 서술}

% \wjb{ActSVD와의 직접 비교를 넣을 필요까지는 없다고 생각함. 오히려 ActSVD의 표현보다, "comparison with Null space projection in Original Space"라는 제목으로 비교하면 어떨까 생각함. ActSVD 자체가 우리의 비교 대상은 아니기 때문임.}

% \wjb{이 섹션을 맨 마지막으로 내리면 어떨까 싶음.}

A common way to separate safety-relevant and task-relevant components is to construct subspaces in the original parameterization and explicitly project weights or updates onto, or away from, these subspaces. In particular, null-space projection can remove components aligned with a utility subspace from a safety-related subspace, yielding a more safety-specific component. WSR-Tune differs in that it does not directly project the original weight matrix onto a predefined safety subspace. Instead, it first constructs a safety-conditioned coordinate system and performs importance estimation and selective protection within that reparameterized space.

Activation-based subspace analysis~\citep{assessing_the_brittleness} provides a representative example of this projection-based approach: it derives safety- and utility-related subspaces from projected activations and isolates safety-specific components by projecting the safety subspace onto the null space of the utility subspace.

\paragraph{Original-space projection.}
Let $W_l\in\mathbb{R}^{d_{\mathrm{out}}\times d_{\mathrm{in}}}$ denote a linear projection and $X_l\in\mathbb{R}^{d_{\mathrm{in}}\times N}$ its input activations. This approach performs SVD on the projected activations,
\begin{equation}
    W_lX_l \in\mathbb{R}^{d_{\mathrm{out}}\times N},
\end{equation}
and uses the leading output singular vectors
$U_{l,r}\in\mathbb{R}^{d_{\mathrm{out}}\times r}$
to construct
\begin{equation}
    W_l^{c}
    =
    U_{l,r}
    (U_{l,r})^\top W_l.
\end{equation}
It further isolates safety-specific modifications through projections such as
\begin{equation}
    \Delta W_l
    =
    (I-\Pi_l^{u})\Pi_l^{s}W_l.
\end{equation}
Thus, this approach explicitly separates safety-specific components in the original parameterization by first projecting onto the safety subspace and then removing components shared with the utility subspace.

\paragraph{WSR-Tune.}
WSR-Tune instead constructs a complete basis from the covariance of safety input activations,
\begin{equation}
    H_lH_l^\top
    =
    U_l\Sigma_lU_l^\top,
\end{equation}
and reparameterizes the weight matrix as
\begin{equation}
    \widetilde W_l
    =
    V_l^\top W_lU_l.
\end{equation}
Safety importance is then evaluated for each coefficient
$\widetilde W_l(i,j)$, and downstream adaptation is restricted by
\begin{equation}
    \Delta W_l
    =
    V_l
    \left(
        \Delta\widetilde W_l
        \odot (1-M_l)
    \right)
    U_l^\top.
\end{equation}
Therefore, WSR-Tune does not preserve an entire rank component. Its preservation unit is an individual coefficient in the reparameterized basis, and the resulting update is not required to be low-rank.

With the default $V_l=I$, let
\begin{equation}
    S_{l,i}=\{j:M_l(i,j)=1\}
\end{equation}
denote the basis directions protected for output row $i$. The corresponding projector is
\begin{equation}
    P_{l,i}
    =
    \sum_{j\in S_{l,i}}u_ju_j^\top.
\end{equation}
Hence, different output rows can preserve different subsets of the safety-conditioned input basis. In contrast, rank-level projection couples parameters through shared output-rank components. This provides WSR-Tune with a finer-grained preservation mechanism than rank-level projection.

The distinction also applies to importance scoring. Since
\begin{equation}
    \frac{\partial L}{\partial \widetilde W_l}
    =
    V_l^\top
    \frac{\partial L}{\partial W_l}
    U_l,
\end{equation}
coordinate-wise importance ranking depends on the chosen basis. Thus, applying the same gradient-based criterion after reparameterization generally identifies a different protected subspace from applying it directly in the original parameterization.

\begin{table}[]
\centering
\caption{Comparison with rank- and neuron-level fine-tuning defenses on Llama-2-7B-Chat with GSM8K. JB AVG denotes average ASR and downstream performance is GSM8K accuracy. Raw scores are percentages and deltas are percentage-point differences relative to Full Params FT. We use $\Delta_{\mathrm{Overall}}=-\Delta_{\mathrm{S}}+\Delta_{\mathrm{D}}$. The green cell highlights the Overall change of WSR-Tune.}
\label{tab:actsved_comparison}
\vspace{5pt}
\resizebox{0.85\textwidth}{!}{
\begin{tabular}{lccccc}
\toprule
 & \multicolumn{2}{c}{\textbf{Safety $\downarrow$}} & \multicolumn{2}{c}{\textbf{Downstream $\uparrow$}} & \textbf{Overall $\uparrow$} \\
\cmidrule(lr){2-3} \cmidrule(lr){4-5} \cmidrule(lr){6-6}
\textbf{Method} & \textbf{JB AVG} & $\boldsymbol{\Delta_{\mathrm{S}}}$ & \textbf{GSM8K} & $\boldsymbol{\Delta_{\mathrm{D}}}$ & $\boldsymbol{\Delta_{\mathrm{Overall}}}$ \\
\midrule
Rank-level protection
& 15.11 & -5.67 & 39.42 & -1.75 & +3.92 \\
Neuron-level protection
& 15.77 & -5.01 & 40.33 & -0.84 & +4.17 \\
\textbf{WSR-Tune}
& 6.90 & -13.88 & 38.99 & -2.18 & \cellcolor{green!20}\textbf{+11.70} \\
\bottomrule
\end{tabular}
}
\end{table}

\paragraph{Fine-tuning-stage comparison.}
To compare these protection strategies, we adapt the component-identification procedures of~\citep{assessing_the_brittleness} into downstream fine-tuning-stage defenses at both the rank and neuron levels. Rank-level protection preserves the safety-relevant output-rank components identified from
\begin{equation}
    W_0X_{\mathrm{safe}}
    =
    U\Sigma V^\top,
\end{equation}
whereas neuron-level protection freezes the identified safety-critical neurons. Table~\ref{tab:actsved_comparison} reports the results under the same Llama-2-7B-Chat and GSM8K setting as our main experiments.

WSR-Tune substantially improves safety preservation over both original-space protection variants while maintaining comparable downstream performance. These results support the benefit of coefficient-level selection in a safety-conditioned reparameterized space, rather than preserving coupled rank or neuron components in the original parameterization.

\section{Additional Related Work on Harmful Fine-Tuning Defenses}
\label{app:additional_related_work}
We complement Section~\ref{sec:related} with a broader discussion of harmful fine-tuning defenses, emphasizing parameter preservation and update constraints. Following the stage-based taxonomy of \citep{huang2024harmful}, we distinguish alignment-stage preparation, interventions during downstream fine-tuning, and post-tuning restoration. The mechanism-based groups below are not mutually exclusive and do not assume identical threat models or access to safety data.

\paragraph{Safety-critical components and parameter freezing.} Safety-related components have been investigated at different granularities. \citep{li2025safety} identify middle layers important for safety and propose SPPFT, which freezes these layers while updating the remaining parameters. \citep{assessing_the_brittleness} examine safety-critical neurons and low-rank components, including activation-derived safety and utility projections in their low-rank analysis. Their analysis also shows that freezing identified safety-critical neurons does not generally prevent subsequent fine-tuning attacks. These findings distinguish identifying components responsible for existing safety behavior from ensuring its preservation throughout adaptation.

\paragraph{Optimization and gradient constraints.} Several fine-tuning-stage methods control how parameters change rather than freezing entire layers. AsFT~\citep{asft} uses the weight difference between aligned and unaligned models as an alignment direction and penalizes update components orthogonal to it, motivated by a narrow safety basin. Lisa~\citep{lisa} alternates safety and task optimization, using proximal penalties relative to the preceding switching checkpoint to limit drift. SafeGrad~\citep{yi2026safegrad} removes task-gradient components that conflict with an alignment gradient before combining the two gradients; its alignment objective uses KL divergence to a frozen safety-aligned reference model. SPF~\citep{zhang2026understanding} instead extracts a low-rank subspace from safety-loss gradients and, when safety and utility gradients conflict, projects the utility gradient onto its orthogonal complement. Non-conflicting updates are retained. These methods therefore differ in the safety signal used to constrain adaptation: checkpoint differences, optimization proximity, reference distributions, or gradient-derived subspaces.

%\wjb{SafeGrad를 추가한다면 reference에 추가하는 것이 필요.}

\paragraph{Low-rank and continual adaptation.} SaLoRA~\citep{salora} combines task-specific adapter initialization with a fixed projection that restricts adapter-induced feature changes to directions orthogonal to estimated safety features. GuardSpace~\citep{zhang2026guardrail} uses covariance-preconditioned weight decomposition to initialize adapters from low-sensitivity components and additionally projects adapter updates into a null space derived from harmful-prompt activations. Its initialization and null-space restriction are separate components. SafeAnchor~\citep{guo2026safeanchor} addresses cumulative safety erosion across sequential domain adaptations: it estimates a Fisher-based safety subspace in LoRA parameter space, incrementally updates this subspace, and combines gradient projection with safety-triggered corrective replay. Thus, low-rank parameterization alone does not specify a common defense mechanism; feature preservation, null-space restrictions, and continual safety monitoring impose different constraints.

\paragraph{Safety supervision and data selection.} SafeInstr~\citep{safeInstr} augments downstream training with safety examples. SEAL~\citep{seal} instead learns sample weights through bilevel optimization, evaluating the safety loss of a model trained on weighted downstream data, and then selects high-ranked downstream samples for fine-tuning. SPARD~\citep{chen2026spard} selects a compact safety dataset using a relevance--diversity determinantal point process. After each utility update, it applies a trust-region-limited correction based on a linearized safety constraint when the safety loss exceeds a threshold. SEAL selects downstream training samples, whereas SPARD selects safety examples used to constrain optimization.

\paragraph{Token- and attention-level constraints.} \citep{constrained_sft} show that refusal alignment can be concentrated in early response tokens and investigate both deeper safety alignment and constrained fine-tuning. Their constrained objective limits deviations from the aligned model with token-dependent regularization, especially at the beginning of a response. Surgery~\citep{liu2026surgery} operates on attention behavior, penalizing positive attention-sink divergence measured between harmful and safe responses to harmful prompts. These methods constrain output or attention behavior during optimization rather than directly freezing a selected block of weight coefficients.

\paragraph{Perturbation and representation-based alignment.} Alignment-stage defenses seek to make subsequent harmful adaptation less effective. Vaccine~\citep{huang2024vaccine} optimizes alignment under adversarial perturbations of hidden embeddings. Targeted Vaccine~\citep{liu2025targeted} concentrates perturbation and training on selected layers according to their importance, rather than perturbing all layers uniformly. Representation Noising~\citep{rosati2024representation} combines benign-behavior retention with suppression of harmful outputs and noise-distribution matching of harmful intermediate representations. Booster~\citep{huang2025booster} penalizes the reduction in harmful loss after a simulated harmful parameter perturbation while optimizing the alignment objective. These approaches modify the model's preparation for subsequent adaptation, rather than necessarily changing the downstream optimizer.

\paragraph{Conditional degradation under malicious adaptation.} Other defenses deliberately couple harmful adaptation with loss of model capability. Self-destructive language models~\citep{wang2026self} encourage opposing gradients on benign and harmful data and incorporate harmful-knowledge unlearning, aiming to make harmful fine-tuning degrade general utility. CTRAP~\citep{yi2026ctrap} plants a collapse objective evaluated after a simulated harmful update, encouraging context-independent fixed-token outputs when the trap is triggered. SDD~\citep{chen2025sdd} trains on harmful instructions with unrelated, high-quality benign responses, so that malicious fine-tuning is intended to impair general capabilities. These methods seek to retain benign usability while accepting capability degradation under malicious adaptation as part of the defense.

\paragraph{Conditioning and model locking.} A related but broader perspective changes the difficulty of subsequent optimization. \citep{zheng2025model} study model immunization through the condition number of the downstream linear-probing Hessian, shaping representations to impede a designated harmful task while preserving pre-training performance. Their linear-probing analysis and vision experiments provide an optimization perspective rather than direct evidence of safety preservation in aligned LLMs. Spectral Deformation~\citep{rosati2025locking} enlarges leading singular values of weight matrices while optimizing a retention loss, aiming to make further training impractically slow without substantially degrading the original task. This model-locking objective restricts subsequent trainability more broadly than safety-preserving task adaptation. 

\paragraph{Post-tuning restoration.} Post-tuning methods modify the result of adaptation rather than its training trajectory. SafeLoRA~\citep{safe-lora} applies alignment-aware projection to learned low-rank updates, while SafeDelta~\citep{safe-delta} adjusts task deltas with safety-aware compensation and RESTA~\citep{Resta} adds a safety vector through task arithmetic. These interventions differ in timing from constraints enforced throughout fine-tuning. 

\paragraph{Relation to WSR-Tune.} WSR-Tune belongs to the fine-tuning-stage family and shares the goal of protecting safety-relevant structure while permitting task adaptation. Its distinction is the combination of a fixed, complete basis derived from safety activations with importance-based selection and protection of reparameterized coefficients. Unlike online gradient-conflict correction, the basis and mask remain fixed during downstream training. Coefficient freezing in this basis induces row-specific linear constraints in the original weights, so it is related to projection-based methods rather than mathematically opposed to them. WSR-LoRA extends this principle through input-factor reparameterization and importance masks on both adapter factors, without equating factor-wise protection with exact freezing of their product. Our contribution is this reparameterization--selection composition, not a new freezing primitive. Detailed geometric and PEFT formulations are provided in Appendices~\ref{app:reparam_analysis} and~\ref{app:wsr_lora}.

%%%%%%%%%%%%%%%%%%%%%%%%%%%%%%%%%%%%%%%%%%%%%%%%%%%%%%%%%%%%
% \clearpage
% \newpage
% \input{checklist.tex}

\end{document}